\documentclass[journal]{IEEEtran}

\usepackage{cite}

\ifCLASSINFOpdf
  \usepackage[pdftex]{graphicx}
  \DeclareGraphicsExtensions{.pdf,.jpeg,.png}
\else
  \usepackage[dvips]{graphicx}
  \graphicspath{{../eps/}}
  \DeclareGraphicsExtensions{.eps}
\fi

\usepackage[svgnames]{xcolor}
\usepackage[colorlinks=true, citecolor=Blue, urlcolor=Blue, linkcolor=Blue]{hyperref}
\usepackage{url}

\usepackage{xcolor}
\usepackage{etoolbox}

\newtoggle{showrevisions}
\togglefalse{showrevisions} 
\newcommand{\revised}[1]{%
  \iftoggle{showrevisions}{{\color{cyan}#1}}{#1}%
}

\newtoggle{showrevisions2}
\togglefalse{showrevisions2} 
\newcommand{\revisedagain}[1]{%
  \iftoggle{showrevisions2}{{\color{cyan}#1}}{#1}%
}

\usepackage{booktabs}
\usepackage{tabularx}

\usepackage{amsmath, amsfonts}
\usepackage{amssymb}
\usepackage{nicefrac}

\ifCLASSOPTIONcompsoc
 \usepackage[caption=false,font=normalsize,labelfont=sf,textfont=sf]{subfig}
\else
 \usepackage[caption=false,font=footnotesize]{subfig}
\fi

\usepackage{url}
\renewcommand{\url}[1]{} 

\usepackage{enumitem}

\usepackage{xcolor}
\usepackage{soul}

\begin{document}

\bstctlcite{IEEEexample:BSTcontrol}

\title{Multiparameter Uncertainty Mapping in Quantitative Molecular MRI using a Physics-Structured Variational Autoencoder (PS-VAE)}

\author{Alex~Finkelstein, Ron~Moneta, Or~Zohar, Michal~Rivlin, Moritz~Zaiss, Shelly~Gur~Zeldich, Dafna~Ben~Bashat, Dinorah~Friedmann-Morvinski, and~Or~Perlman,~\IEEEmembership{Member,~IEEE}
\thanks{A.F., M.R., and O.P. are with the School of Biomedical Engineering, Tel-Aviv University, Tel-Aviv, Israel. O.Z. and D.F.M are with the School of Biochemistry, Neurobiology, and Biophysics, Tel-Aviv University, Tel-Aviv, Israel. M.Z. is with the Institute of Neuroradiology, University Hospital Erlangen, and the Department of Artificial Intelligence in Biomedical Engineering, Friedrich-Alexander Universität Erlangen-Nürnberg (FAU), Erlangen, Germany. S.G.Z and D.B.B are with the Sagol Brain Institute, Tel Aviv Sourasky University Medical Center, Tel Aviv, Israel, and the Gray Faculty of Medical and Health Sciences, Tel Aviv University, Tel Aviv, Israel. D.F.M., R.M., D.B.B, and O.P. are also with the Sagol School of Neuroscience, Tel-Aviv University, Tel-Aviv, Israel. e-mail: orperlman@tauex.tau.ac.il}}

\markboth{IEEE Transactions On Medical Imaging}%
{Shell \MakeLowercase{\textit{et al.}}: Bare Demo of IEEEtran.cls for IEEE Journals}

\maketitle

\begin{abstract}

Quantitative imaging methods, such as magnetic resonance fingerprinting (MRF), aim to extract interpretable pathology biomarkers by estimating biophysical tissue parameters from signal evolutions. However, the pattern-matching algorithms or neural networks used in such inverse problems often lack principled uncertainty quantification, which limits the trustworthiness and transparency, required for clinical acceptance.
Here, we describe a physics-structured variational autoencoder (PS-VAE) designed for rapid extraction of voxelwise multi-parameter posterior distributions. Our approach integrates a differentiable spin physics simulator with self-supervised learning, and provides a full covariance that captures the inter-parameter correlations in the latent biophysical space. The method was validated in a multi-proton pool chemical exchange saturation transfer (CEST) and semisolid magnetization transfer (MT) molecular MRF study, across in-vitro phantoms, tumor-bearing mice, healthy human volunteers, and subjects with glioblastoma and Alzheimer's Disease. The resulting multi-parametric posteriors are in good agreement with those calculated using a brute-force Bayesian analysis, while providing an orders-of-magnitude acceleration in whole brain quantification.
In addition, we demonstrate how monitoring the multi-parameter posterior dynamics across progressively acquired signals provides practical insights for protocol optimization and may facilitate real-time adaptive acquisition. The code and sample data are  available at \href{https://github.com/momentum-laboratory/ps-vae}{https://github.com/momentum-laboratory/ps-vae}.

\end{abstract}


\IEEEpeerreviewmaketitle

\section{Introduction}
\IEEEPARstart{S}{aturation} transfer magnetic resonance imaging (ST-MRI) is an increasingly investigated molecular imaging technique \cite{zaiss2022theory} that has already provided biological insights across a variety of preclinical\cite{zhou2017optimized, sun2007detection, pavuluri2019noninvasive} and clinical\cite{jones2018clinical, zhou2022review, vinogradov2023cest} applications. ST mechanisms comprise several contrast pathways, including chemical exchange saturation transfer (CEST), relayed nuclear Overhauser enhancement (rNOE), and semisolid magnetization transfer (MT)\cite{zaiss2022theory}.

The CEST-weighted contrast reflects the combination of the proton volume fraction and exchange rate, and is influenced by a variety of confounding factors\cite{zaiss2013chemical}, including water relaxivity, direct water saturation, aliphatic rNOE, and macromolecule-associated signals. As for many other MRI contrast mechanisms, these dynamics have motivated the pursuit of a fully quantitative method, capable of isolating and mapping the biophysical parameters of interest.

In consequence, quantitative CEST studies have gradually shifted focus from traditional steady-state signal model fitting\cite{mcmahon_quantifying_2006, woessner_numerical_2005}, to quasi-steady-state\cite{kim2022demonstration}, and pseudo-random acquisition patterns\cite{perlman_quantitative_2022} that enable a shorter acquisition time. Two emerging technologies that play a key role in these endeavors are magnetic resonance fingerprinting (MRF)\cite{ma2013magnetic, cohen_rapid_2018, vladimirov2025quantitative}, and AI-based quantification\cite{chen2020vivo, finkelstein_multi-parameter_2025, kang2021unsupervised}.

Unfortunately, applying these techniques clinically is still hindered by (i) the absence of ground truth in vivo, (ii) the black-box nature of the deep learning models used to date, and (iii) the substantial variability in the biophysical parameter estimates obtained by different algorithmic variants and research groups\cite{liu_quantitative_2013, heo_quantifying_2019, perlman_quantitative_2022, carradus_measuring_2023, finkelstein_multi-parameter_2025}. It is therefore essential to characterize the uncertainties and biases of each estimated parameter pixel-wise, to foster trust in the AI-based quantitative imaging pipeline. 

Our goal was to develop a rapid computational method to assess the joint
uncertainties of quantitative MRI parameter estimates, by approximating the full posterior within a real-time neural-network-based quantification framework. The method was validated across various quantitative molecular (ST) MRI tasks, where the lack of appropriate multi-parameter uncertainty estimates constitutes an urgent need; however, the same concept can be applied for a variety of other quantitative imaging contrasts. 

\begin{figure*}[!t]
\centering
\includegraphics[width=0.99\textwidth]{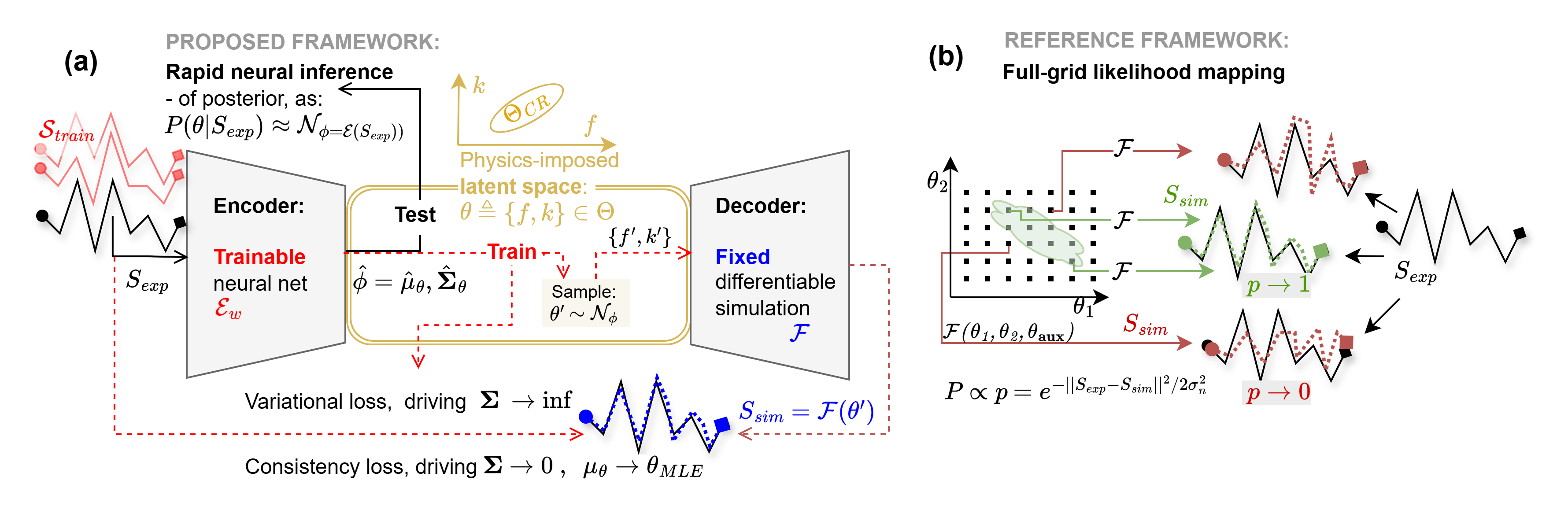}
\caption{\textbf{Computational pipelines for multi-parameter posterior estimation in quantitative molecular MRI}. \textbf{(a)} Rapid uncertainty estimation using a physics-structured variational autoencoder (PS-VAE). A multi-layer perceptron (\(\mathcal{E}_w\), red) is trained using self-supervised learning on experimentally measured signals \(S_{m}=S_{exp} \in \mathcal{S}_{train}\). At inference, voxelwise parameterization \(\hat{\phi}({S_{m}},w)=\hat{\mu},\hat{\boldsymbol{\Sigma}}=\mathcal{E}_w(S_{exp})\) of an approximate posterior \( \hat{P}(\theta|S_{exp}) \approx Q_{\hat{\phi}} = \mathcal{N}(\hat{\boldsymbol{\mu}}, \hat{ \boldsymbol{\Sigma} })  \) is obtained. The derived confidence region (CR) \(\Theta_{CR} \subset \Theta\) in the latent space of the tissue parameters \(\Theta\), reflects consistency with the measured signal under a biophysical model (\( \mathcal{F} \), blue), so that  \(\mathcal{F}(\theta' \in \Theta_{CR}) \approx S_{exp} \). The 
CR is an oriented uncertainty ellipsoid around the mean estimate \(\hat{\boldsymbol{\mu}}\), represented using a non-diagonal covariance matrix \(\hat{\boldsymbol{\Sigma}}\). A random posterior sample $\theta' \sim Q_{\hat{\phi}}$ is decoded at each iteration during training, and a $\boldsymbol{\Sigma}$-regularizing term balances the cycle-consistency loss.
\
\textbf{(b)} Reference framework for likelihood-mapping. Each experimentally measured signal \(S_m=S_{exp}\), associated with a given voxel, is compared (by means of dot-product) to a local dictionary of signals across a dense grid in the parameter space $\Theta =\Theta_1 \times \Theta_2$. The discrepancy map is then translated into a likelihood and, consequently, into the posterior.}
\label{fig:frameworks}
\end{figure*}

\subsection{Related Work } 
The standard formalization of seeking uncertainty-aware solutions of inverse problems such as MRI reconstruction and quantification is the Bayesian posterior estimation. The poor scaling of non-parametric approaches motivated functional approximations known as variational inference (VI)\cite{blei_variational_2017} or variational Bayes (VB)\cite{attias_variational_1999, chappell_variational_2009}. The variational autoencoder (VAE)\cite{kingma_auto-encoding_2022} provided further scaling via data-driven \textit{amortization}, altering the inference stage from solving a per-observation optimization problem to a single forward pass of a shared neural network (NN). 
A separate line of work augmenting generic auto-encoding (AE) focuses on guidance by biophysical modeling. This recent concept concerns physics-informed self-supervised learning \cite{liu_magnetic_2021, jun_ssl-qalas_2023, huang_uncertainty-aware_2022, minore_resolving_2025 ,finkelstein_multi-parameter_2025}, where the AE latent is associated with the point estimate of the physical parameters. This estimate is decoded by simulation into a 'denoised' output signal whose discrepancy with the experimental measurements then guides the training in order to improve the resilience to realistic signal distortions\cite{finkelstein_multi-parameter_2025}. Here, we integrated these two approaches (probabilistic- and physics-driven AE extensions) within a simple yet meaningful multi-parametric framework. 

The concept of using a VAE over physics-grounded latent space has been explored in various contexts such as VAE-based perfusion MRI\cite{zhang_bayesian_2024} and spatiotemporal modeling\cite{glyn-davies_-dvae_2024, goh_solving_2021} among others, with ref. \cite{goh_solving_2021} discussing applicability to uncertainty quantification (UQ) in inverse problems. However, these studies relied on synthetic datasets for training, incorporating heuristic assumptions on data and error distributions. In addition, previous VI methods apply the mean-field approximation, and neglect parameter correlations under the posterior, in seeking to predict a diagonal covariance matrix for the Gaussian approximation. Recent works extended VAE to use a full or structured covariance, e.g. in the context of image formation and compressed sensing\cite{duff_vaes_2023}. Similar ideas have been used to predict multivariate posteriors over dynamic contrast enhanced (DCE) MR parameters \cite{bliesener_efficient_2020}, but the reported training used synthetic data in a supervised learning fashion. The latter setting admits simpler techniques, such as mean-variance estimating (MVE) networks \cite{nix_estimating_1994}, which have recently been trained on synthetic Z-spectra to assess univariate uncertainties in a semi-quantitative CEST parameter estimation task (accelerating standard Lorentzian fitting) \cite{glang_deepcest_2020}.  

\revised{Other related physics-aware UQ approaches include the conditional VAE (CVAE) \cite{gabbard_bayesian_2022, tonolini_variational_2020, liu_posterior_2023, zhang_cvae-based_2025}, which utilizes a conventional abstract latent space to represent the solution manifold, with physics entering indirectly via synthetic training data. In this framework, posterior sampling is achieved by performing multiple passes through the decoder, each using a random latent sample conditioned on the observation. This sampling process is procedurally similar to Bayesian neural networks (BNNs) implemented via Monte-Carlo dropout \cite{gal_dropout_2016} or deep ensembles \cite{lakshminarayanan_simple_2017, wilson_bayesian_2022}. Both BNNs and CVAEs require explicit ground truth for training, limiting them to synthetic regimes and risking adverse effects of model misspecification when transferred to real data \cite{wang2025deep, finkelstein_multi-parameter_2025}. While CVAE can represent arbitrary posteriors, simpler alternatives, such as MVE networks, can suffice for posteriors well approximated by a Gaussian \cite{nix_estimating_1994}. }
Here, we introduce an amortized variational framework that explicitly captures the multivariate posterior structure within a self-supervised regime, enabling a rapid and robust probabilistic quantitative imaging. 
\revised{
Specifically, we redesign the MVE-net concept to accommodate the absence of ground-truth samples, treating it as a VAE encoder while moving beyond the mean-field approximation to capture latent correlations. We train this full-covariance MVE network directly on measured signals in a self-supervised manner, enforcing cycle consistency by sampling from the predicted posterior and propagating the samples through a differentiable forward simulator (simulation-in-the-loop). The resulting net can be used for real-time approximate Bayesian quantification of unseen scans of the same protocol and similar tissues. Furthermore, as a second operational mode resembling the 'untrained networks' approach \cite{qayyum2022untrained, heckel_deep_2024}, the network's outputs after fitting a single subject can yield potentially optimal instance-specific posteriors.
}

\subsection{Main contributions}
\begin{enumerate}[label=(\alph*)]

    \item \textit{Full-covariance quantification}: robust estimation of the joint posterior, capturing inter-parameter correlations.
    \item \textit{Physics-structured VAE architecture}: using the forward physical model as the decoder, we learn probabilistic inversion without synthetic data or ground-truth labels.
    \item \textit{Versatility}: robust performance across instance-specific and amortized VI modes: (i) fitting a single observation, and (ii) transfer to rapid inference on unseen subjects.
    \item \textit{Proof-of-concept applications}: (i) UQ-aware analysis and discovery of multiparameter biomarkers (ii) Posterior extraction from partial measurements, facilitating protocol optimization and real-time adaptive acquisition.
\end{enumerate}

\begin{figure*}[!t]
\centering
\includegraphics[width=0.99\textwidth]{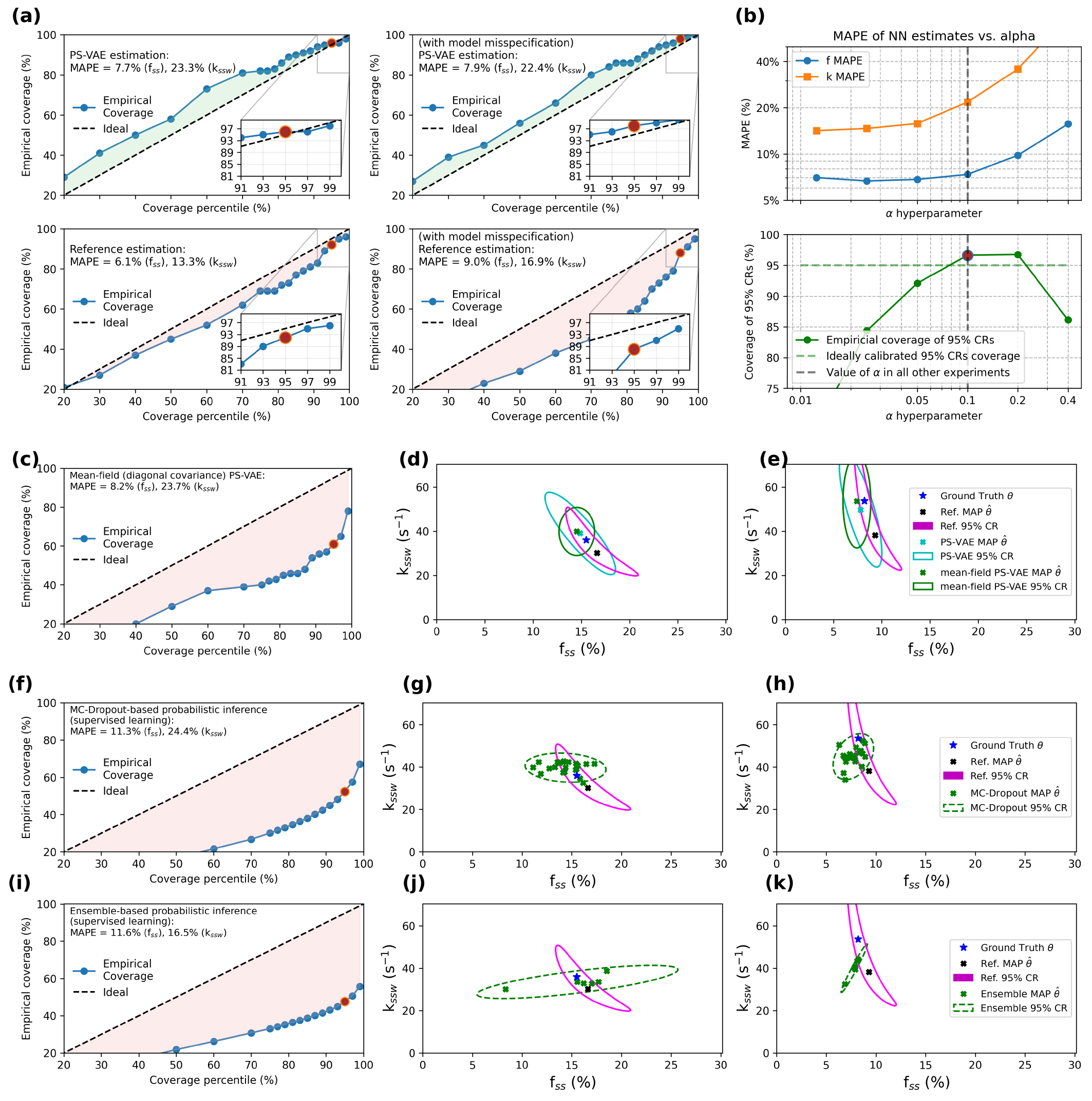}
\caption{\revised{\textbf{Calibration and ablation studies using a digital phantom}. \textbf{(a)} Posterior calibration curves of frequentist coverage probability for PS-VAE (top) and reference method (bottom), for fully specified (left) and misspecified - non-AWGN residuals (right) forward model of physics and measurements used for signal simulation. \textbf{(b)} The impact of the training hyperparameter $\alpha$ on the estimation accuracy (top) and UQ coverage (bottom). The dashed vertical line marks the value $\alpha$=0.1 used throughout this work, balancing the MAPE performance (top) with CR coverage (bottom). \textbf{(c-k)} Ablation studies of restricting PS-VAE to diagonal covariance (c-e), and replacing PS-VAE by conventional supervised learning on synthetic data extended for UQ by MC-dropout (f-h) or deep ensemble (i-k) techniques. In each row, a calibration curve on the left (c,f,i) is followed on the right by parameter space maps for the same two representative simulated voxels (d-e, g-h, j-k) on which the estimates and confidence regions (CRs) provided by the method-under-test (green) are compared with the ground truth (blue) and the reference method CR (magenta). For the alternative deep UQ methods, the dots correspond to multiple inferences  (each yielding a point prediction), using the same network with random dropout realizations (g-h) or differently trained ensemble members (j-k). The CR for the figure (dashed curve) is constructed from Gaussian modeling of those points. 
}}\label{fig:simulated}
\end{figure*}

\begin{figure*}[!t]
\centering
\includegraphics[width=0.9\textwidth]{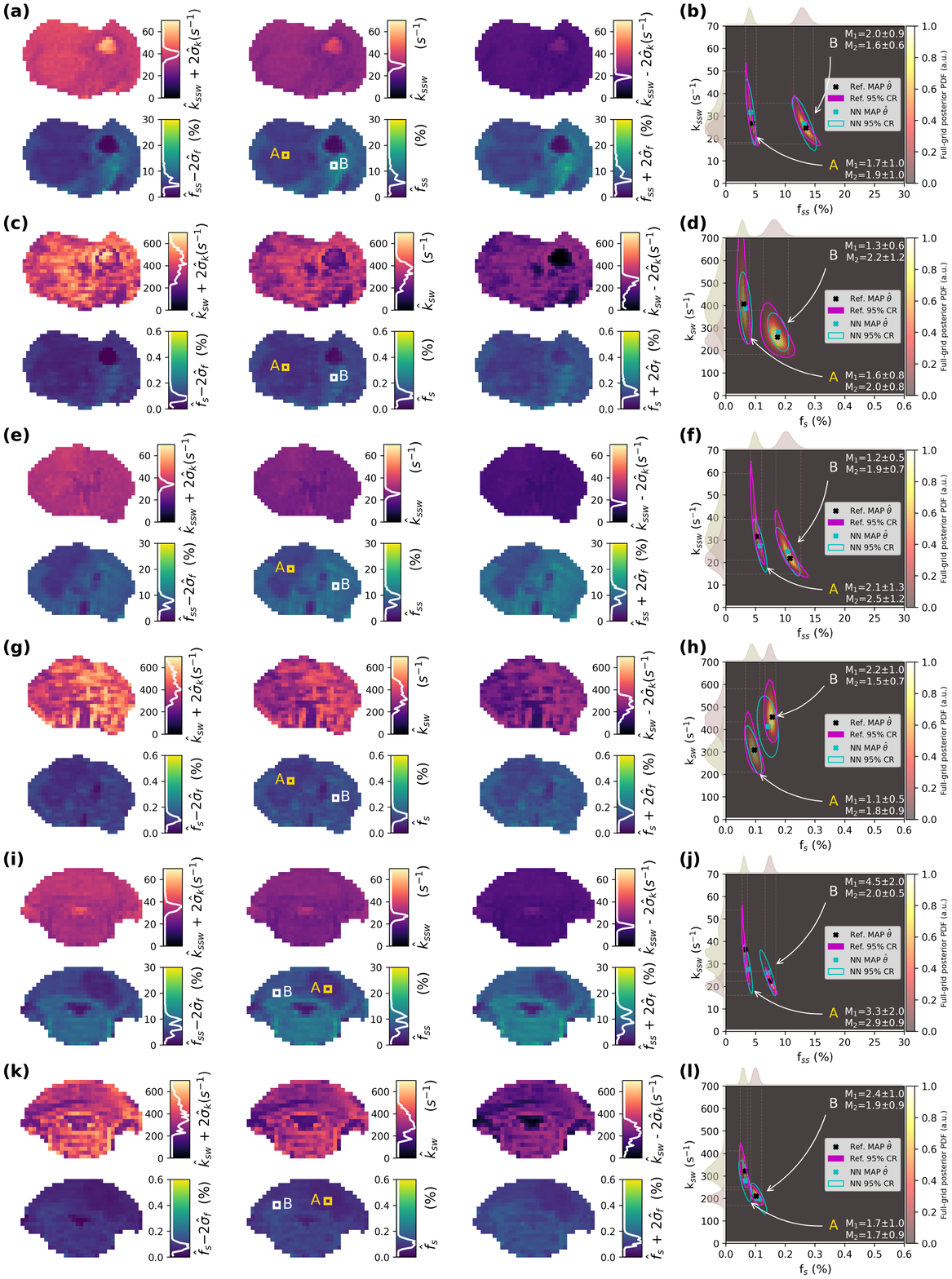} 

\caption{\textbf{Uncertainty-aware quantification of the semisolid MT (a-b, e-f, i-j) and amide (c-d, g-h, k-l) proton exchange rates (top) and volume fractions (bottom) in three (out of n=7) representative tumor-bearing mice.} Panel groups a-d, e-h, i-l refer to each of the three mice respectively. The color-coded brain images (a,c,e,g,i,k) show the PS-VAE output maps of maximum-a-posteriori (MAP) estimates of tissue parameters (center) and the corresponding univariate confidence interval (CI) bounds; the distributions across the entire slice are displayed inside the colorbar. Multi-parametric confidence regions (CR) are provided alongside each parameter map (b,d,f,h,j,l) for randomly chosen pixels in the ipsitumoral and contralateral regions of interest (ROIs), marked by "A" and "B" within the MAP images. The CRs were obtained from Gaussian posteriors given by PS-VAE output (cyan) or from the reference grid-based method (free-form CR bounds in magenta, underlying posteriors as heatmaps). The agreement between the two posteriors as quantified using the Mahalanobis distance is provided as text within the figure. \revisedagain{These visual elements are used in the same way in Figs. \ref{fig:vol_mtcest_maps},\ref{fig:patient_MT},\ref{fig:ablations_in_vivo} below.}
}
\label{fig:mouse_maps}
\end{figure*}

\section{Methods}

\subsection{Problem statement: Bayesian inversion of multi-pool MRI}
\revised{Multi-pool saturation MRI dynamics can be described by the Bloch-McConnell (BM) ordinary differential equations (ODE), with the general form $\dot{M}=AM+C$, where $M_{3p}$ is the concatenation of 3D rotating-frame magnetization vectors of the $p$ proton pools. Here, p=3 accounting for the free water, semisolid MT, and amide protons. The ODE coefficients, as aggregated in the system matrix $A_{3p\times3p}$, are derived from the tissue parameters, e.g., the proton volume fraction (f$_s$) and exchange rate (k$_{sw}$) of each compound pool,} 
and are directly affected by a variety of acquisition parameters, including the saturation pulse power ($B_1$) and frequency offset ($\Delta\omega$). For ST-MRF, the imaging scheme is repeated with pseudo-randomly varied saturation pulse parameters\cite{vladimirov2025quantitative}, essentially sampling the non-steady-state magnetization signal. 
The inverse problem of quantitative molecular MRI requires inferring the ODE coefficients from samples at different time points. 
Existing schemes for rapid NN-based parameter quantification can be formalized as a prediction of the maximum-likelihood point estimate (MLE) of biophysical parameters \(\theta\) from a measured signal \(S_m\), under an assumed biophysical model \(\mathcal{F}\), so that \(\mathcal{F}(\theta_{MLE}) \approx S_m\) \cite{cohen_rapid_2018}. Here, we extended this concept by approximating a Bayesian estimate of the joint posterior distribution \(P(\mathbf{\theta}|S_m)\). Furthermore, while classical MRF inherently relies on synthetic signals, we employ amortization over real (experimentally acquired) data, to avoid distribution shift concerns. 

\subsection{Proposed framework: rapid quantification of the posterior}
We extend the self-supervised architecture\cite{finkelstein_multi-parameter_2025}, interpreted as a physics-structured autoencoder (PS-AE)\cite{minore_resolving_2025}, to a variational version, termed PS-VAE (Fig. \ref{fig:frameworks}a). The true distribution \(P(\mathbf{\theta}|S_m)\) is approximated by (i) designing the NN encoder to specify a member \(\phi\) of a Gaussian parametric family \( Q_{\phi} = \mathcal{N}(\boldsymbol{\mu}, \boldsymbol{\Sigma}) \), (ii) sampling to back-propagate through probabilistic prediction ("reparameterization trick"\cite{kingma_auto-encoding_2022}), and (iii) adding a regularizing term that balances the consistency loss. In contrast to vanilla VAEs, we impose a fixed decoder using the biophysical model \(\mathcal{F}\), which aligns the latent space to that of the tissue parameters \(\boldsymbol{\theta}\in\Theta\). \revisedagain{This perspective motivates the adoption of the AE/VAE terminology, even as the decoder is not a learnable neural network, to conceptualize the inverse problem of quantification as compactly representing measured data via a meaningful latent \cite{minore_resolving_2025, goh_solving_2021}.} 
Distribution flexibility is provided by predicting a complete (non-diagonal) covariance matrix \(\boldsymbol{\Sigma}\), corresponding to an "uncertainty ellipsoid" around the mean estimate \( \boldsymbol{\mu} \). The weights \( w \) of the multi-layer perceptron (MLP) encoder \( \mathcal{E}_w \) are learned in a self-supervised manner from a training set \( \mathcal{S}_{\text{train}} \) comprised of experimentally measured signals. For each raw signal \( S_m \), the encoder outputs the mean vector \( \boldsymbol{\mu} \) and the covariance matrix \( \boldsymbol{\Sigma}\) for a multivariate Gaussian, representing the posterior distribution of the latent tissue parameter vector \( \boldsymbol{\theta} \). The VAE-inspired learning objective aims at uncertainty-regularized fitting:
\begin{equation}
\min_{w} \; \mathbb{E}_{S_m \in \mathcal{S}_{\text{train}},\theta' \sim \hat{Q}(S_m)} \left[ 
\mathcal{L}_c\left(S_m, \boldsymbol{\theta}'\right)
+ \mathcal{L}_{\text{reg}}(\boldsymbol{\mu}, \boldsymbol{\Sigma})
\right],
\label{Eq:PS-VAE-loss}
\end{equation}
where \( \boldsymbol{\hat{\phi}} = \boldsymbol{\mu}, \boldsymbol{\Sigma} = \mathcal{E}_w(S_m) \), \( \boldsymbol{\theta}' \sim Q_{\hat{\boldsymbol{\phi}}}=\mathcal{N}(\boldsymbol{\mu}, \boldsymbol{\Sigma}) \) is a posterior sample per step. The first term is the consistency loss, penalizing the residual between the measured and simulated signals:
\begin{equation}
\mathcal{L}_c(S_m, \boldsymbol{\theta}') = \left\| S_m - \mathcal{F}(\boldsymbol{\theta}') \right\|_2^2
\end{equation}
where \( \mathcal{F} \) is the differentiable biophysics-based simulator. The second term is the VAE regularization, as informed by the Kullback-Leibler (KL) divergence of posterior from a prior - an origin-centered $\sigma$=1 Gaussian in standard VAEs, towards an analytical form: \( L_{KL}=\Sigma_j\left(\mu_j^2+\sigma_j^2-\log \sigma_j\right)\). 
\revisedagain{Here, we retain the last term, adapting it for a full covariance, to reflect a neutral uniform prior across the physics-structured latent space and focus on addressing posterior covariance collapse:} 
\begin{equation}
    L_{reg}=-\nicefrac{\alpha}{2}\log \det \boldsymbol{\Sigma}=-\alpha\log \det \boldsymbol{S}=-\alpha\Sigma_j\log \lambda_j
\end{equation} 
where \(\alpha\) is a hyperparameter and the diagonal \( \boldsymbol{S} = \boldsymbol{\lambda I}\) refers to the result of the eigen-decomposition \( \boldsymbol{\Sigma} = \mathbf{U} \mathbf{S}^2 \mathbf{U}^\top \). The latter is also utilized in the re-parametrization trick sampling, for drawing \( \theta' \sim \mathcal{N}(\boldsymbol{\mu}, \boldsymbol{\Sigma}) \) from standard normal samples via:
\begin{equation}
\boldsymbol{\theta}' = \boldsymbol{\mu} + \mathbf{U} \mathbf{S} \boldsymbol{\epsilon}, \quad 
\boldsymbol{\epsilon} \sim \mathcal{N}(\boldsymbol{0}, \mathbf{I}_d),
\end{equation}
In practice, the non-redundant terms of diagonal $\mathbf{S}$ and orthonormal \( \mathbf{U} \) matrices are drawn from the encoder output.
\revised{
In the 2D (f,k) case, this amounts to the two eigenvalues of $\mathbf{\Sigma}$ and the rotation angle, comprising together with $\boldsymbol{\mu}$ the five parameters defining the location, dimensions, and orientation of the elliptical Confidence Region (CR).

To improve model accuracy and reduce the complexity of the multi-parameter space, non-proton-exchange parameters were measured externally and accounted for voxel-wise in a sequential manner, as previously performed in \cite{finkelstein_multi-parameter_2025, perlman_quantitative_2022}. Specifically, the auxiliary parameter vector comprised of field corrections and water pool parameters, \(\boldsymbol{\theta}^{(1)}_{aux}=\{T_1, T_2, B_0, B_1\}\), was used to improve the ("stage 1") estimation of the semisolid MT parameters, \( \boldsymbol{\theta}_{MT}=\{f_{ss}, k_{ssw}\}\), from measurements modeled by $\mathcal{F}_{MT}(\boldsymbol{\theta}^{(1)}_{aux},\boldsymbol{\theta}_{MT})$. Next, the auxiliary parameter vector comprised of field corrections and water and semisolid pool parameters, \(\boldsymbol{\theta}^{(2)}_{aux}=\{T_1,T_2,B_0,B_1, \hat{f}_{ss}, \hat{k}_{ssw} \}\), was used to estimate ("stage 2") the quantitative APT parameters, \( \boldsymbol{\theta}_{APT}=\{f_{s}, k_{sw}\}\), from the corresponding protocol modeled by $\mathcal{F}_{APT}(\boldsymbol{\theta}^{(2)}_{aux},\boldsymbol{\theta}_{APT})$. In both stages, the auxiliary parameter vector was passed to both the encoder ($\phi^{v}=\mathcal{E}(\boldsymbol{\theta}^{v},\boldsymbol{\theta_{aux}}^{v})$ at voxel $v$) and the decoder $\mathcal{F}$ as used in either the PS-VAE training/fitting (see above) or in the reference method (see next section).
}

\begin{figure*}[!t]
\centering
\includegraphics[width=0.95\textwidth]{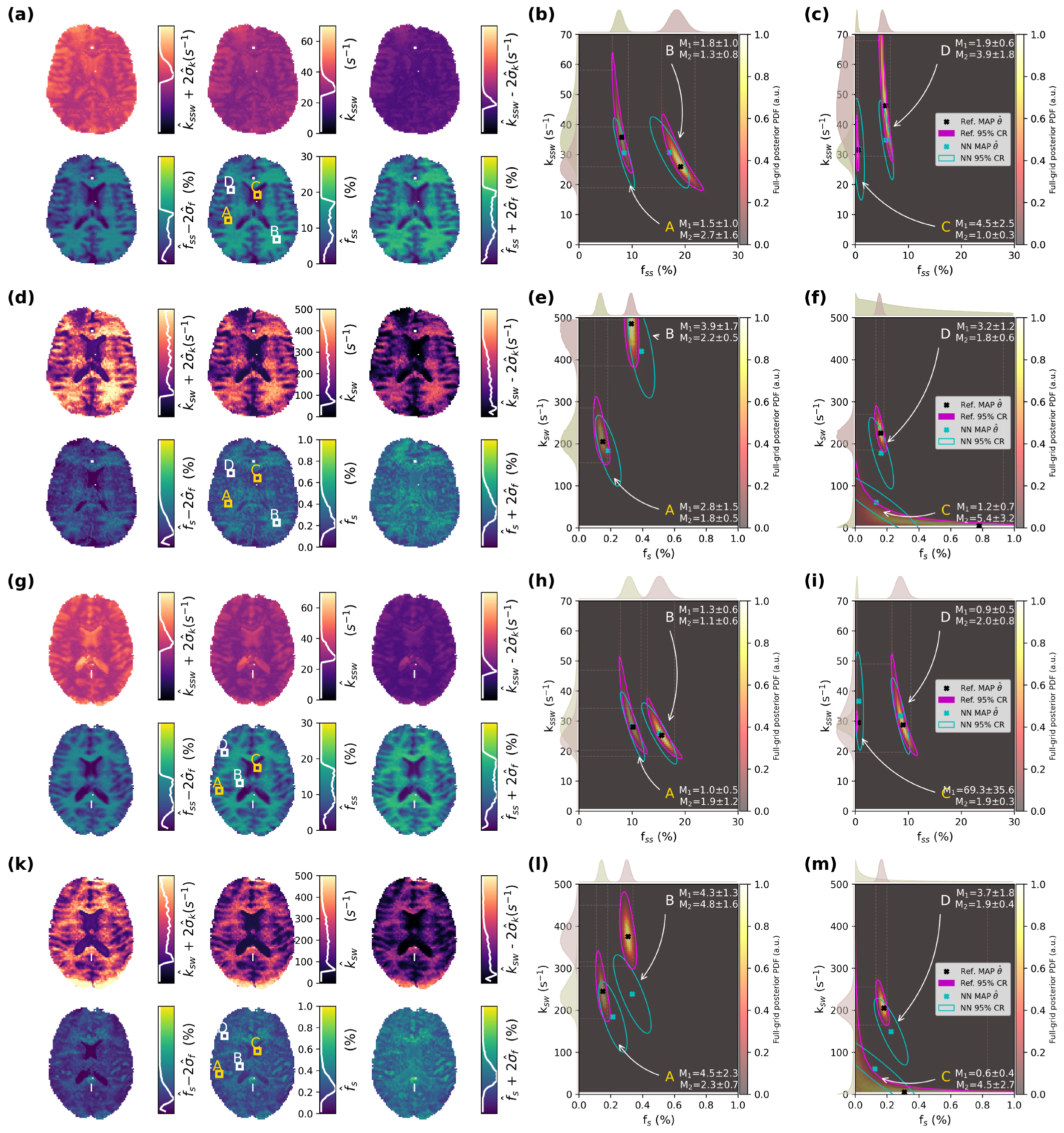} 
\caption{\textbf{Uncertainty-aware quantification of the  semisolid MT (a-c, g-i) and \revised{amide (d-f, k-m)} proton exchange rates (top) and volume fractions (bottom) in two of the n=4 healthy human volunteers.} Panels a-f (g-m) represent a single axial slice out of the first (second) volunteer 3D brain scan. The color-coded brain images (a,d,g,k) show the PS-VAE output of uncertainty-aware maps.
Multi-parametric confidence regions (CR) are provided (b-c,e-f,h-i,l-m) in each row alongside the output parameter maps group, contrasting pairs of randomly chosen pixels in the white and gray matter regions of interest (ROIs) in (b,e,h,i) marked by "A" and "B" within the MAP images, and similarly pairs of ventricle vs. gray-matter voxels marked by "C" and "D" in (c,f,i,m). 
\revisedagain{Uncertainty visualization components follow the conventions established in Fig. \ref{fig:mouse_maps}}.
The per-subject PS-VAE maps were obtained using a network trained on other subjects (as described in Sec. \ref{sec:loocv-transfer-eval}). 
} \label{fig:vol_mtcest_maps}
\end{figure*}

\subsection{Reference framework: likelihood mapping}
\label{sec:reference_framework}
In the absence of ground truth for in vivo likelihood mapping, we provide a straightforward reference method that follows the true posterior as closely as possible. For a signal \(S_m\) at a given voxel, the Bayesian posterior is described by the likelihood and prior as: 
\begin{equation}
P(\boldsymbol{\theta}|S_m)=\frac{f(\boldsymbol{\theta})}{\int_{\boldsymbol{\theta}' \in \Theta}f(\boldsymbol{\theta}')} ,\ f(\boldsymbol{\theta})=P(\boldsymbol{\theta})P(S_m|\boldsymbol{\theta})
\end{equation}
The need to compute \(f(\boldsymbol{\theta})\) across the entire support \( \Theta\) for the denominator, typically discourages the use of this straightforward approach. However, for a case of low-dimensional estimation (e.g., \(\boldsymbol{\theta} \in \mathbb{R}^2\)), \revised{with all other ("auxiliary") parameters previously captured as $\boldsymbol{\theta}_{aux}$, }a brute force computation across a reasonably detailed grid is both feasible (at least for a small number of voxels) and beneficial, for a transparent comparison. We assume the same biophysical model and an additive white Gaussian noise (AWGN) for the residuals of the \(n\) samples comprising the signal: 
\begin{equation}
S_m=(S^{(j)})_{j=1}^n \sim \mathcal{N}\left(\mathcal{F}(\boldsymbol{\theta}_{true}, \mathbf{\theta}_{aux}), \sigma_m  \boldsymbol{I}_{n\times n} \right)
\label{Eq:s_m}
\end{equation}

We then simulate a dedicated dictionary of synthetic signals across a dense grid $\Theta _{d}= \Delta \theta_{1}\mathbb{Z}\times \Delta \theta_{2}\mathbb{Z}\cap \Theta$ for a given \(S_m\), \(\boldsymbol{\theta}_{aux}\) at a voxel of interest: 
\begin{equation}
\{S_{sim}(\boldsymbol{\theta})|\boldsymbol{\theta} \in \Theta_d\} = \{ \mathcal{F}(\theta_{1,2}, \boldsymbol{\theta}_{aux}) | \theta_{1,2} \in  \Delta \theta_{1,2}\mathbb{Z}_N\} 
\label{Eq:gengrid}\end{equation}
As a next step, the error model is plugged in for explicit computation of the likelihood map across the parameters grid: 
\begin{align}
 P(S_{m}| \boldsymbol{\theta}, \mathcal{F}, \sigma_{meas})&\propto e^{-\sum_{j=1}^n (\mathcal{F}(\boldsymbol{\theta})[j]-S_{m}[j])^2/(2\sigma^2)}
\label{eq:err2prob}      
\end{align}
The posterior is generated by multiplying the likelihood by a prior, and then normalizing to obtain a proper probability distribution function (PDF) $P(\boldsymbol{\theta}|S_{m})$. 

\subsection{Confidence region definition}
The bi-variate 95\% confidence regions (CRs) are defined for each estimation method, as the iso-posterior contours enclosing 95\% of the probability density. For Gaussian posteriors (as provided by PS-VAE), the CRs reduce to Mahalanobis ellipses \((\mathbf{x}-\boldsymbol{\mu})^\top\!\boldsymbol{\Sigma}^{-1}(\mathbf{x}-\boldsymbol{\mu})=\chi^2_{2,0.95}\) (chi-square: \(\chi^2_{2,0.95}\!\approx\!5.99\)),
with semi-axes \(a,b=\sqrt{\chi^2_{2,0.95}\lambda_{1,2}}\) and orientation
\(\phi=\arctan(v_{1y}/v_{1x})\), where \(\lambda_{1,2}\) and \(\mathbf{v}_{1,2}\) are the eigenvalues and the corresponding eigenvectors of \(\boldsymbol{\Sigma}\) in descending order. The univariate confidence intervals (CIs) were similarly defined using marginalized posteriors; namely, using the $\pm2\sigma$ obtained from the diagonal of $\boldsymbol{\Sigma}$, for the Gaussian case.

\subsection{Evaluation criteria and performance metrics} 
\label{sec:eval-metrics}

\subsubsection{Goodness of fit for point estimates} 
This criterion spatially maps the voxelwise fitting errors $||\mathcal{F}(\hat{\boldsymbol{\theta}})-S_m||$ (as normalized mean squared error, NRMSE) of the neural network maximum a posteriori (MAP) estimate $\hat{\boldsymbol{\theta}}=(\hat{f},\hat{k})=\boldsymbol{\mu}_{\theta}(S_m)$\ compared to the value provided by the MAP estimate of the exact Bayesian $min(||\mathcal{F}(\boldsymbol{\theta})-S_m||)=||\mathcal{F}(\boldsymbol{\theta}_{MAP})-S_m||$. In addition to reflecting the consistency 
between the raw experimental data and the model-based "re-synthesized" signals, this criterion also provides a rough 
estimate of model uncertainty.

\subsubsection{Univariate confidence interval (CI) accuracy}
The crude agreement of the UQ-aware univariate estimates as made by PS-VAE and the reference method was quantified as the percentage of pixels with any intersection between the parameter CIs provided by the two methods. 

\subsubsection{Rigorous quantification of the multi-parameter posterior agreement}
\label{mahalanobis_explanation}
The ability to capture the non-trivial joint parameter distributions was evaluated by using the 2-D Mahalanobis distance $\mathcal{M}(\theta,\mathcal{D})$ to compare the posteriors obtained by  PS-VAE and by full-grid likelihood-mapping. The metric was calculated in a two-directional manner, namely as distances between samples from the distribution $P_1(\theta|s)$ provided by one method to distribution $P_2(\theta|s)$ provided by the other, and vice versa: $\{\mathcal{M}(\theta', P_1)|\theta' \sim P_2\}, \{\mathcal{M}(\theta', P_2)|\theta' \sim P_1\}$. The empirical distribution of these values was then compared to the ideal PDF curve of the Mahalanobis distances between samples and the distribution from which they had been drawn $\mathcal{M}(\theta_{[\sim P]},P)\sim\chi_{d=2}$ (with equality achieved at $P_1=P_2$). A threshold of M=4 was determined as a "marginal agreement", heuristically extending the CI-intersection criterion $\theta^{(1)}+2\sigma^{(1)}>\theta^{(2)}-2\sigma^{(2)}$ to a multivariate similarity.

\begin{figure*}[!t]
\centering
\includegraphics[width=0.98\textwidth]{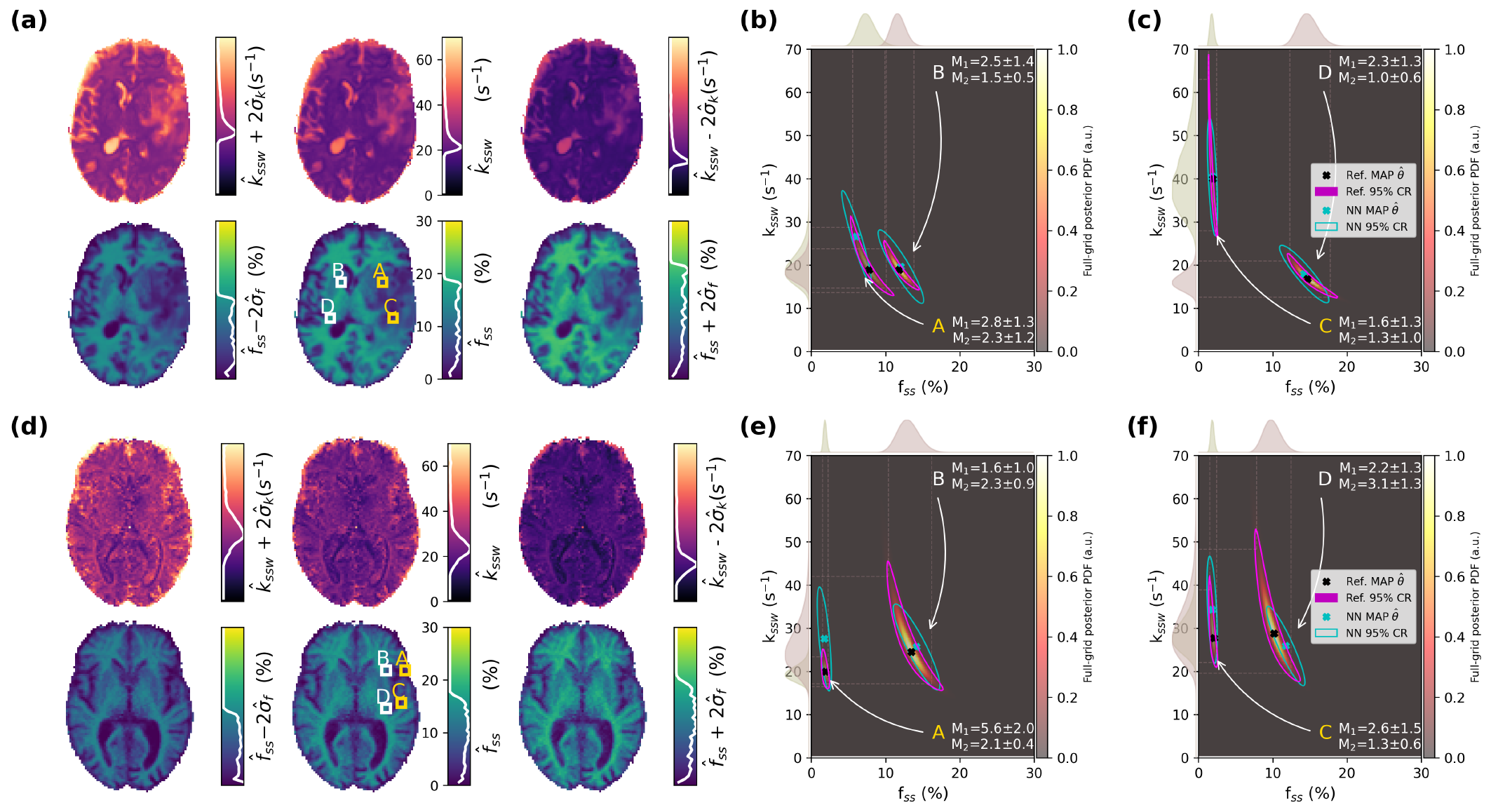} 
\caption{\textbf{Uncertainty-aware quantification of the semisolid MT proton exchange rates (a,d, top) and volume fractions (a,d, bottom) in patients with glioblastoma (a-c) \revised{and Alzheimer's Disease (d-f)}.}  The color-coded brain images (a,d) represent the PS-VAE output uncertainty-aware maps. 
Multi-parametric confidence regions (CR) are provided in (b,c,e,f) for randomly chosen pixels in the ipsitumoral, contralateral regions of interest (ROIs), marked by 'A','B','C', and 'D' (for the subject with glioblastoma), or representative gray matter ('A','C') and white matter ('B','D') pixels for the subject with Alzheimer's Disease. 
\revisedagain{Uncertainty visualization components follow the conventions established in Fig. \ref{fig:mouse_maps}}.
}
\label{fig:patient_MT}
\end{figure*}

\begin{figure*}[!t]
\centering
\includegraphics[trim=0 0 0 0, clip, width=0.99\textwidth]{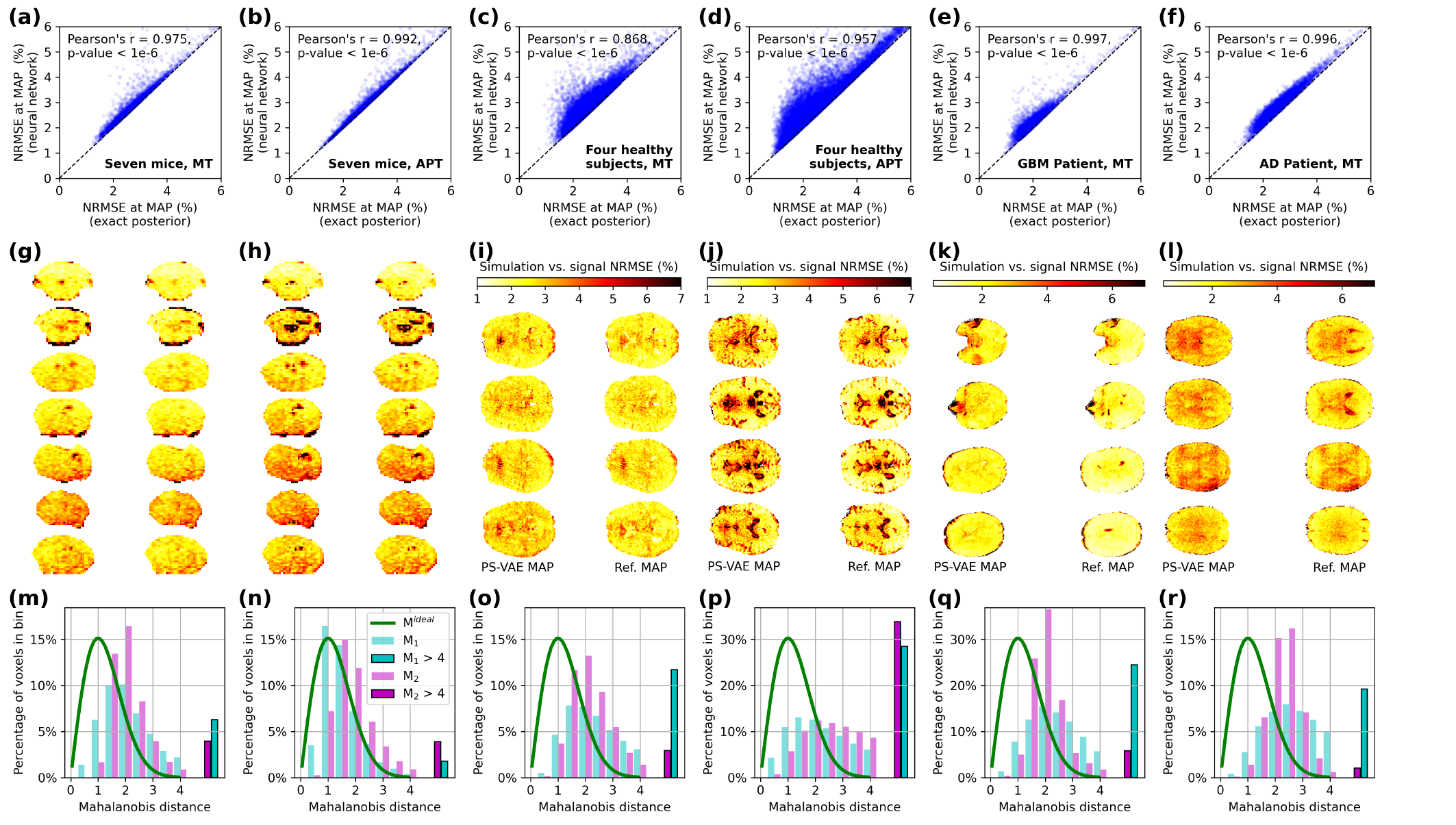} 
\caption{\textbf{Quantitative performance evaluation and statistical overview of the PS-VAE-derived output with respect to the reference method.} \textbf{(a-f)}. Correlation plots comparing the NRMSE (reflecting the goodness of fit to signal) of the maximum-a-posteriori (MAP) parameter values obtained by PS-VAE (y-axis) and the reference Bayesian quantification method (x-axis), for all voxels of quantification of semisolid MT in mice (a), APT in mice (b), semisolid MT in healthy volunteers (c) APT in healthy volunteers (d) semisolid MT in a patient with glioblastoma (e), \revised{and Alzheimer's Disease (f)}. \textbf{(g-l)}. NRMSE maps for PS-VAE-derived vs. reference method-derived tissue parameters, with panels corresponding to the same list of setups as in (a-f). In (g-h), the rows present the single slice of each of the 7 mice; in (i-j) one slice per each human volunteer; in (k-l), multiple slices of respective subjects. \textbf{(m-r)}. Complete distributions of the Mahalanobis distances of samples from PS-VAE-derived posteriors to the reference posteriors (and vice-versa), aggregated from all voxels, with panels referring to the same list of setups as (a-f).
The ideal distribution (of Mahalanobis distances of samples from any distribution $P$ to same $P$) is shown by the green curves. 
} \label{fig:aggregated_stats}
\end{figure*}

\section{Experiments}

\revised{
\subsection{Digital phantom study}
\label{sec:digital_phantom_design}

A digital phantom was constructed to evaluate the UQ performance of the methods in a controlled scenario. We synthesized 100,000 voxels with randomly chosen parameter combinations (via Latin Hypercube Sampling) for f$_{ss}$, k$_{ssw}$, and T$_1$. T$_2$, B$_0$, B$_1$ in the ranges 5-25\%, 10-60 s$^{-1}$, 800-1800 ms, 50-120 ms, -0.3-0.3 ppm, and 90-110\%, respectively. For each virtual voxel, the 'measured signal' has been created by simulating (i.e., applying the physical model $\mathcal{F}$, see Sec. \ref{sec:reference_framework}) the response associated with the clinical MT-MRF protocol (see Sec. \ref{sec:mri_acquisition_protocols}). Next, a Gaussian noise of a randomly chosen magnitude in the range 1-4\% (40-28 dB SNR) was added \cite{perlman_cest_2020}. After using this data for training, we tested the resulting posterior predictors on 500 new and different data points. These were drawn from the same distribution, while optionally adding a random bias of 1-4\% to one third of the measurements, thus simulating a model misspecification. The known ground truth (GT) was used to compute coverage levels - the fraction of cases in which the GT parameter vector lies within the X\% CR - as a function of the nominal level X, yielding the calibration curve. The process was repeated with/without the model misspecification, for PS-VAE and the reference method (likelihood mapping), and then across several ablation experiments: (i) Changing the loss term weighting (hyperparameter $\alpha$, see Fig. \ref{fig:simulated}b). (ii) PS-VAE with diagonal covariance restrictions (Fig. \ref{fig:simulated}c-e). (iii) Alternative trained approaches using supervised learning of probabilistic networks (Fig. \ref{fig:simulated}f-k). Alongside UQ performance, the accuracy of the MAP point estimate with respect to the ground truth has been quantified using the mean absolute percentage error (MAPE) metric, for each of the parameters f$_{ss}$, $k_{ssw}$. The alternative Bayesian quantification networks were trained on the digital phantom data using the GT as a target (corresponding to the standard accelerated-fingerprinting approach \cite{cohen_mr_2018, perlman_quantitative_2022, vladimirov2025quantitative}). To provide UQ in this setting, we implemented two of the common probabilistic-NN techniques, namely Deep Ensembles and MC-Dropout, on top of the supervised training. To calibrate the uncertainty intervals to a physically meaningful order of magnitude, we performed a parameter sweep across the baseline variants. This settled on a 0.4 dropout probability for MC-dropout, and a variable early-stopping strategy spanning 200–1200 steps as the diversity-generating mechanism for the deep ensemble \cite{qendro2021early}.
}
\subsection{Experimental datasets}

\subsubsection{L-arginine (L-arg) phantoms at 9.4 T} Open access data from an in-vitro MRF study in L-arg phantoms was used  (\href{https://doi.org/10.6084/m9.figshare.23289800}{https://doi.org/10.6084/m9.figshare.23289800})\cite{nagar_dynamic_2023}. The phantoms consisted of three glass vials containing 50 mM L-arg with the pH titrated to 5.0, 5.5, and 6.0.

\subsubsection{Tumor-bearing mice at 7T} All experimental protocols adhered to the ethical principles of the Israel National Research Council (NRC) and received approval from the Tel Aviv University Institutional Animal Care and Use Committee (IACUC). Two cell lines (O1, a cell line derived from a lentiviral-induced pediatric glioma\cite{rousso2021p32}, and GP01, a murine diffuse midline glioma primary tumor cell line\cite{ross2024microglia}) were used to establish tumor mouse models in seven mice. All the mice were imaged using a 7T scanner (Bruker, Germany). 

\subsubsection{Healthy human volunteers at 3T}
Four healthy volunteers (three females/one male, with an average age of 23.75$\pm$0.83) were scanned at Tel Aviv University using a 3T MRI (Prisma, Siemens Healthineers). The research protocol was approved by the TAU Institutional Ethics Board (study no. 2640007572-2) and the Chaim Sheba Medical Center Ethics Committee (0621-23-SMC). 

\subsubsection{A subject with brain cancer at 3T} A patient with 
Glioblastoma was scanned with a 3T MRI (Prisma, Siemens Healthineers) at the University Hospital Erlangen (FAU) \cite{nagar2025multi}, using the same imaging protocol as for healthy volunteers. The research protocol was approved by the University Hospital Erlangen Institutional Review Board and Ethics Committee. The participant provided written, informed consent before the study. 

\revised{
\subsubsection{A subject with Alzheimer's Disease at 3T}
A patient with Alzheimer's Disease was scanned using a 3T MRI (Cima.X, Siemens Healthineers) at the Tel Aviv Sourasky University Medical Center. The research protocol was approved by the Institutional Review Board and Ethics Committee (study no. 0038-08-TLV). The participant provided written informed consent before the study.
}

\subsection{MRI acquisition protocols}
\label{sec:mri_acquisition_protocols}
A continuous-wave (CW) MRF protocol with a single-slice spin-echo echo-planar-imaging (SE-EPI) readout was used for both phantoms and mice, with a saturation pulse frequency offset fixed at 3 ppm (phantoms), 3.5 ppm (APT imaging), or varied between 6-14 ppm (semisolid MT imaging)\cite{cohen_rapid_2018, perlman_quantitative_2022}. 
T$_1$, T$_2$, B$_0$, and B$_1$ maps were acquired using rapid acquisition with relaxation enhancement (RARE), multi-echo spin-echo (MSME), water saturation shift referencing (WASSR)\cite{kim2009water}, and the double angle protocols, respectively. The field of view (FOV) was set to 32$\times$32 mm / 19$\times$19 mm and the slice thickness to 5 mm / 1.5 mm for phantoms/mice, respectively. The matrix size was set to 64$\times$64 pixels for both phantoms and mice. For humans, T$_1$, T$_2$, B$_0$, and B$_1$ maps were obtained using saturation recovery, multi-echo, and the WASABI\cite{schuenke2017simultaneous} sequences, respectively. All human protocols used a \revised{Pulseq-CEST hybrid sequence} \cite{herz2021pulseq} \revised{playing out the ST parameter encoding preparation PW sequence followed by a} 3D snapshot readout with whole-brain coverage\cite{mueller2020whole} and an isotropic resolution of 1.8 mm\cite{finkelstein_multi-parameter_2025}, \revised{with a single exception for the subject with Alzheimer's disease, who was scanned with a 2 mm $\times$ 2 mm $\times$ 5 mm resolution covering 12 slices} \cite{deshmane20193d}. A spin-lock saturation train (13$\times$100 ms, 50\% duty-cycle) was used for ST. The complete sequence definitions are listed in \cite{vladimirov2025quantitative}.

\vspace{-0.1em}
\subsection{Computational framework implementation details}
The framework was implemented in JAX\cite{jax2018github}  on a desktop computer equipped with an NVIDIA GeForce RTX 3060 GPU. A simple voxelwise multi-layer perceptron (MLP) with two hidden layers of 256 neurons each has been employed as the PS-VAE encoder. The training used batches comprised of 10 image slices, a learning rate (LR) of 0.001 in the clinical MT imaging case, with 350 epochs when training on 3 human volunteers and 1000 epochs when fitting a single patient volume. \revised{The training for clinical APT imaging used batches of 1/4 slice (via downsampling, to meet memory constraints) and 9 epochs, with LR=0.001}. For preclinical (both MT-CW and APT-CW) imaging, 1-slice batches, LR=0.01 and 1000 epochs have been used. 
Complete code and sample data are available at \href{https://github.com/momentum-laboratory/ps-vae}{https://github.com/momentum-laboratory/ps-vae}. 

\subsection{Operating modes and evaluation procedures}
\label{sec:loocv-transfer-eval}
\subsubsection{Rapid Bayesian quantification of unseen data}
Applying the PS-VAE-trained rapid neural posterior estimator to held-out data is referred to as \textit{transfer mode} and can be seen as a manifestation of \textit{amortized variational inference}. This has been evaluated on semisolid MT data from healthy volunteers, using a leave-one-out cross-validation (LOOCV) routine: training was performed on n-1=3 subjects (using a total of 135 slices) and tested on the held-out data from the n$^{th}$ subject. The process was repeated across all n=4 split options.
\subsubsection{Subject-specific training and inference}
\label{sec:small-data training}
The subject-specific ('test-time') training and transfer-less (same-subject) inference of PS-VAE are referred to as the \textit{fitting mode}. \revised{This can be seen as approaching \textit{instance-specific variational inference}, albeit amortized across voxels of a subject sharing the network's weights} \cite{finkelstein_multi-parameter_2025}. This mode was applied to all phantom and mouse data, as well as to the clinical human scans in subjects with Glioblastoma \revised{and Alzheimer's Disease.}

\begin{table*}[t!]
\renewcommand{\arraystretch}{1.1}
\caption{Performance Evaluation}  
\label{table:agreement}
\centering
\scriptsize

\begin{tabular}{|lllllll|}
\toprule
Metric / target \& setup & TB mice (n=7) & TB mice (n=7) & HV\ humans (n=4) & HV\ humans (n=4) & GBM\revised{, AD} patients\ & Phantom (n=3 vials) \\
  & APT\ 7T\ CW & \ ssMT\ 7T\ CW  & \ ssMT\ 3T\ PW &  APT\ 3T\ PW & ssMT\ 3T\ PW  & L-arginine 9.4T CW \\
\midrule
NRMSE\ for MAP values & NN: 3.3\%$\pm$0.7\%  & NN: 3.0\%$\pm$0.3\%  & NN: 2.9\%$\pm$0.1\%  & NN: 3.2\%$\pm$0.1\%  & NN: 2.4\%, 3.1\% & NN: 2.2\%$\pm$0.0\%  \\
("goodness of fit") & Ref: 3.2\%$\pm$0.6\%  & Ref: 2.8\%$\pm$0.3\%  & Ref: 2.6\%$\pm$0.1\%  & Ref: 2.8\%$\pm$0.03\% & Ref: 2.2\%, 2.8\% & Ref: 2.1\%$\pm$0.0\%  \\
\hline
Univariate\ CIs\ intersect & f$_{s}$:\ 99.9\%$\pm$0.1\%\  & f$_{ss}$:\ 99.4\%$\pm$0.5\%\  & f$_{ss}$:\ 98.5\%$\pm$0.5\%\  & f$_{s}$:\ 98.0\%$\pm$1.3\% & f$_{ss}$:\ 97.8\%, 100\%,  & f$_{s}$:\ 100.0\%$\pm$0.0\% \\
(percentage of voxels) & k$_{sw}$:\ 99.5\%$\pm$0.4\%\  & k$_{ssw}$:\ 98.2\%$\pm$1.0\%\  & k$_{ssw}$:\ 97.1\%$\pm$1.0\%\  & k$_{sw}$:\ 97.8\%$\pm$1.5\% & k$_{ssw}$:\ 96.6\%, 99.9\% & k$_{sw}$:\ 100.0\%$\pm$0.0\% \\
\hline
Mahalanobis\ distance,\  & M$_1$=1.41$\pm$0.13\  & M$_1$=2.16$\pm$0.40\  & M$_1$=2.53$\pm$0.61\  & M$_1$=2.69$\pm$0.27 & M$_1$=2.69, 2.82 & M$_1$=2.01$\pm$0.38 \\
(median\ across\ voxels) & M$_2$=1.87$\pm$0.24\  & M$_2$=2.02$\pm$0.11\  & M$_2$=2.07$\pm$0.32\  & M$_2$=3.18$\pm$0.36 & M$_2$=1.98, 2.3 & M$_2$=1.57$\pm$0.10 \\
\hline
Posteriors discrepancy & M$_1$\textgreater 4:\ 3.7\%$\pm$1.4\%  & M$_1$\textgreater 4:\ 12.9\%$\pm$6\%& M$_1$\textgreater 4:\ 20.2\%$\pm$11\%\  & M$_1$\textgreater 4:\ 28.4\%$\pm$6\%  & M$_1$\textgreater 4:\ 24.5\%, 19.2\% & M$_1$\textgreater 4:\ 0.4\%$\pm$0.6\% \\
(percentage\ of\ voxels) & M$_2$\textgreater 4:\ 8.1\%$\pm$5.6\%  & M$_2$\textgreater 4:\ 8.1\%$\pm$4.0\%  & M$_2$\textgreater 4:\ 5.2\%$\pm$1.2\%\  & M$_2$\textgreater 4:\ 33.8\%$\pm$8\%  & M$_2$\textgreater 4:\ 5.8\%, 2.1\% & M$_2$\textgreater 4:\ 0.0\%$\pm$0.0\%  \\
\hline
\revised{Posterior divergence metrics & KL:\ 1.19$\pm$0.53\  & KL:\ 1.74$\pm$0.23\  & KL: 2.19$\pm$0.77\  & KL:\ 4.96$\pm$1.25 & KL: 2.03, 2.66 & KL:\ 0.93$\pm$0.32 \\
(median\ across\ voxels) & JS:\ 0.26$\pm$0.07\  & JS:\ 0.45$\pm$0.06\  & JS: 0.53$\pm$0.14\  & JS:\ 0.66$\pm$0.05 & JS: 0.57, 0.64 & JS:\ 0.30$\pm$0.10 \\
 & HL:\ 0.44$\pm$0.06\  & HL:\ 0.61$\pm$0.05\  & HL: 0.66$\pm$0.10\  & HL:\ 0.76$\pm$0.03 & HL: 0.70, 0.74 & HL:\ 0.48$\pm$0.09 \\
 }
\bottomrule
\end{tabular}
\flushleft{HV: healthy volunteer; TB: tumor-bearing; APT: amide proton transfer; ssMT: semisolid magnetization transfer; CW/PW: continuous/pulsed wave; GBM: glioblastoma multiforme; \revised{AD: Alzheimer's Disease; KL: Kullback-Leibler divergence; JS: Jensen-Shannon divergence; HL: Hellinger distance}}.
\end{table*}

\section{Results and discussion}
\subsection{Validation and ablation study in a digital phantom}
\label{sec:digital_phantom_results}
\revised{The digital phantom experiment described in Sec. \ref{sec:digital_phantom_design} was used for objective evaluation of the ability of the posteriors to capture the known ground truth, for the described methods and selected alternatives. Fig. \ref{fig:simulated}a compares the resulting empirical coverage obtained for PS-VAE (top) with that obtained using the reference method (bottom), for both fully specified (left) and misspecified (right) signal models. While the reference method is more accurate and well-calibrated when the AWGN assumption strictly holds (the only case where its output can be deemed a 'ground truth posterior'), its performance naturally deteriorates under model misspecification. In contrast, PS-VAE shows robust coverage for both cases. Next, an ablation study was performed to examine the importance of the full uncertainty covariance matrix. Notably, when the PS-VAE is forced to predict a strictly diagonal covariance, the performance deteriorates, even without misspecification (Fig. \ref{fig:simulated}c). This deterioration arises from the mismatch between the true (slanted) posterior shape and the axis-aligned shapes that a mean-field model can capture, since such a model lacks the biophysically inherent parameter correlation. The instance-level analysis of the CRs compared to the GT is further demonstrated for two representative simulated voxels in Fig. \ref{fig:simulated}d-e. As expected, the 95\% CR comparison yields a substantially better visual agreement between PS-VAE trained to generate a full covariance and the reference method, compared to the diagonal covariance (‘mean-field PS-VAE’). The impact of the training hyperparameter $\alpha$ is explored in Fig. \ref{fig:simulated}b for the misspecified-model case; the value of 0.1 (as used in our work) provides an appropriate balance between UQ coverage and estimation accuracy, while too low or too high values either nullify the probabilistic sampling component or make it detrimental to convergence and accurate prediction, consistent with theory.
As an advanced ablation study of the overall PS-VAE approach, we implemented an alternative concept of approximate Bayesian estimation using probabilistic neural networks,  trained in a supervised way. The results (Fig. \ref{fig:simulated} f-k) show that these approaches, while generally viable and providing uncertainties of the right order of magnitude, struggle to provide good UQ coverage and to capture CR shapes. Notably, these shortcomings arise even before transfer to in-vivo data (see Sec. \ref{ablation_in_vivo}). 
}


\subsection{Uncertainty visualization and biomedical interpretation}
\label{sec:uncertainty-visualization}
Representative univariate confidence interval (CI) maps for tumor-bearing mice, healthy human volunteers, and patients with glioblastoma \revised{or Alzheimer's Disease} are shown in Figs. \ref{fig:mouse_maps}a,c,e,g,i,k, \ref{fig:vol_mtcest_maps}a,d,g,k and \ref{fig:patient_MT}a,d respectively. The maps provide immediate visual insights into the reliability of each quantified tissue parameter, as opposed to a "stand-alone" observation of the MAP estimates ($\hat{f}_{ss}$, $\hat{f}_{s}$, $\hat{k}_{ssw}$, $\hat{k}_{sw}$, center column in the above mentioned sub-figures). Specifically, the CI maps for the semisolid MT proton volume fraction ($f_{ss}$) are relatively tight and narrowly distributed, due to the high concentration of semisolid macromolecules in the brain (especially in the white matter). In contrast, the CI maps for the semisolid proton exchange rate ($k_{ssw}$) exhibit a substantially wider distribution, in agreement with previous reports attesting to the difficulty in accurately determining the value of this challenging physical property\cite{vladimirov2025quantitative, kim2020deep}. Similarly, the spatial CI maps of the amide proton exchange rate in mice ($k_{sw}$, Fig. \ref{fig:mouse_maps}c,g,k) exhibit a wide uncertainty (large $\sigma$), which may explain the wide variability in previous reports \cite{liu_quantitative_2013, heo_quantifying_2019, perlman_quantitative_2022, carradus_measuring_2023, finkelstein_multi-parameter_2025}.  
While the univariate CIs represent the first layer of information, deeper insights are provided by the joint multi-parametric posteriors (Figs. \ref{fig:mouse_maps}b,d,f,h,j,l, \ref{fig:vol_mtcest_maps}b-c,e-f,h-i,l-m, and \ref{fig:patient_MT}b-c,e-f). The eccentricity and inclination of the confidence regions in the semisolid MT quantification in mouse and human brains indicates a negative correlation between $f_{ss}$ and $k_{ssw}$ under the posterior. These findings are consistent with simplified analytical models where the leading-order behavior of the ST effect is proportional to the product of the proton volume fraction and exchange rate \cite{zaiss2022theory}. In addition they provide further evidence for the necessity of full covariance for approximating the exact CRs. 
The posterior maps provide a straightforward visualization of the multi-parameter differences across various tissue types; a glance at these maps for healthy human volunteers immediately conveys the differences in proton volume fraction and exchange rates between white and gray matter (WM/GM). These can be seen in the separation between the respective curvilinear CRs (mostly horizontal, namely better reflected by changes in f$_{ss}$). The overlap of the two univariate PDFs obtained by marginalizing, i.e. projecting the posteriors onto the proton volume fraction and exchange rate axes (Figs. \ref{fig:mouse_maps}b,d,f,h,j,l, \ref{fig:vol_mtcest_maps}b-c,e-f,h-i,l-m, and \ref{fig:patient_MT}b-c,e-f) provides an alternative way to assess the discriminative power of a given biophysical parameter. Specifically, when analyzed at a single-pixel level (as shown in Figs. \ref{fig:mouse_maps}-\ref{fig:patient_MT}), the CRs and their projection on the biophysical parameter axes can be used to determine whether an ipsilateral pixel is substantially different from the contralateral counterpart. For example, Fig. \ref{fig:mouse_maps}d,h,l shows that murine tumor edema consistently decreases the amide proton volume fraction\cite{perlman_quantitative_2022}, leading to non-overlapping projected PDFs for f$_{s}$. On the other hand, while the MAP for the ipsitumoral k$_{sw}$ appears larger than the contralateral region for some mice (Fig. \ref{fig:mouse_maps}c), a careful look at the CR map reveals a non-negligible PDF tail overlap (Fig. \ref{fig:mouse_maps}d). This finding is in agreement with a recent 31P imaging study that reported a less than 0.1 pH units\cite{paech2023whole} increase in intracellular pH in the tumor (associated with the amide proton base catalyzed proton exchange rate increase). Unlike tumor proliferation, cell death is known to be associated with a substantial decrease in pH (and amide k$_{sw}$). This is typically seen following successful cancer treatment\cite{nilsson2006cytosolic, perlman_quantitative_2022}. While our study did not involve treatment, the availability of the CR map may enable the determination of certainty for a more subtle form of cell death that is often seen in late tumor stages (Fig. \ref{fig:mouse_maps}g,h).
Importantly, observing the posterior maps across different voxels further emphasizes the heterogeneity of the tissue parameter values for human glioblastoma (Fig. \ref{fig:patient_MT}b,c), compared to the more homogeneous mouse tumor model (Fig. \ref{fig:mouse_maps}).

\subsection{Uncertainty quantification benchmarking in vivo}
\subsubsection{Comparison to reference method}
Table \ref{table:agreement} presents a summary of method performance across various imaging targets, hardware, and setups. Notably, the subject-mean NRMSE representing the goodness of fit of the PS-VAE-derived point estimates are lower than 4\% across all experiments. In addition, the NRMSE values obtained by the PS-VAE-derived MAP estimates are strongly correlated with those obtained by the reference method (e.g., in Figs. \ref{fig:aggregated_stats}a-c,e, r=0.992, r=0.975, r=0.868, and r=0.997 for APT in mice, MT in mice, healthy volunteers, and a GBM patient, respectively, with p$<$0.001 in all cases). 

\revised{ Basic UQ compatibility is indicated by the overlap between the univariate CIs obtained by PS-VAE and those from the reference method seen in more than 95\% of voxels in each of the studies. Similarly, a basic multi-parametric agreement in a typical voxel is indicated by the two-directional Mahalanobis distances of the PS-VAE-derived and reference posteriors (M$_1$,M$_2$, see Sec.\ref{mahalanobis_explanation}). These, with the exception of the 3T pulsed-wave (PW) saturation APT in humans, had a median M$_1$, M$_2$ below 2.7, 2.3, respectively, across all other imaging targets and setups (Table \ref{table:agreement}). In all of these, the M$_2$ (quantifying divergence of sampled 'reference values' from the PS-VAE CR) was below 4 in more than 90\% of the voxels. 
} 
The full distributions of the Mahalanobis distances, across the various imaging experiments are shown in Fig. \ref{fig:aggregated_stats}m-r. Thirty-six visual examples of posteriors alongside the corresponding Mahalanobis values are provided for a variety of random GM/WM/CSF pixels (Fig. \ref{fig:vol_mtcest_maps}b,c,e,f,h,i,l,m; Fig. \ref{fig:patient_MT}e,f) and tumor/contralateral pixels (Figs. \ref{fig:mouse_maps}b,d,f,h,j,l; \ref{fig:patient_MT}b,c). 


\revised{Apart from the CI-intersect and Mahalanobis-based measures, we also quantified posterior similarity using dedicated distribution-divergence metrics, as described in Table 1. The values of Kullback-Leibler divergence (KL) generally correspond to M$_2$ in terms of overall performance variation across use cases. It should be noted that the Bayesian reference method (likelihood mapping across a grid in the parameter space) is not an exact ground-truth posterior in a strict sense when applied to in-vivo data. It strongly relies on the modeling assumptions and its coverage deteriorates when these assumptions are inaccurate (Fig. 2a). Thus, certain deviations should be tolerated, especially the ‘conservative’ ones where PS-VAE CRs are somewhat larger, covering, in addition to most of the reference CRs, some areas assigned improbable under the (possibly overly narrow) reference posterior.
}
The rapid quantification of unseen data, using the PS-VAE-trained decoder, generated consistent MAP and CI maps (as part of the LOOCV study, Fig. \ref{fig:vol_mtcest_maps}) with a between-subjects coefficient-of-variance (CV) of: CV(f$_{ss}$)=2.5\%/7.2\% and CV(k$_{ssw}$)=2.2\%/2.3\%, for the WM/GM regions, respectively. At the same time, there was consistently good model-data fit, as shown in Fig. \ref{fig:aggregated_stats}c,d,i,j. 

\begin{figure*}[!t]
\centering
\includegraphics[width=0.99\textwidth]{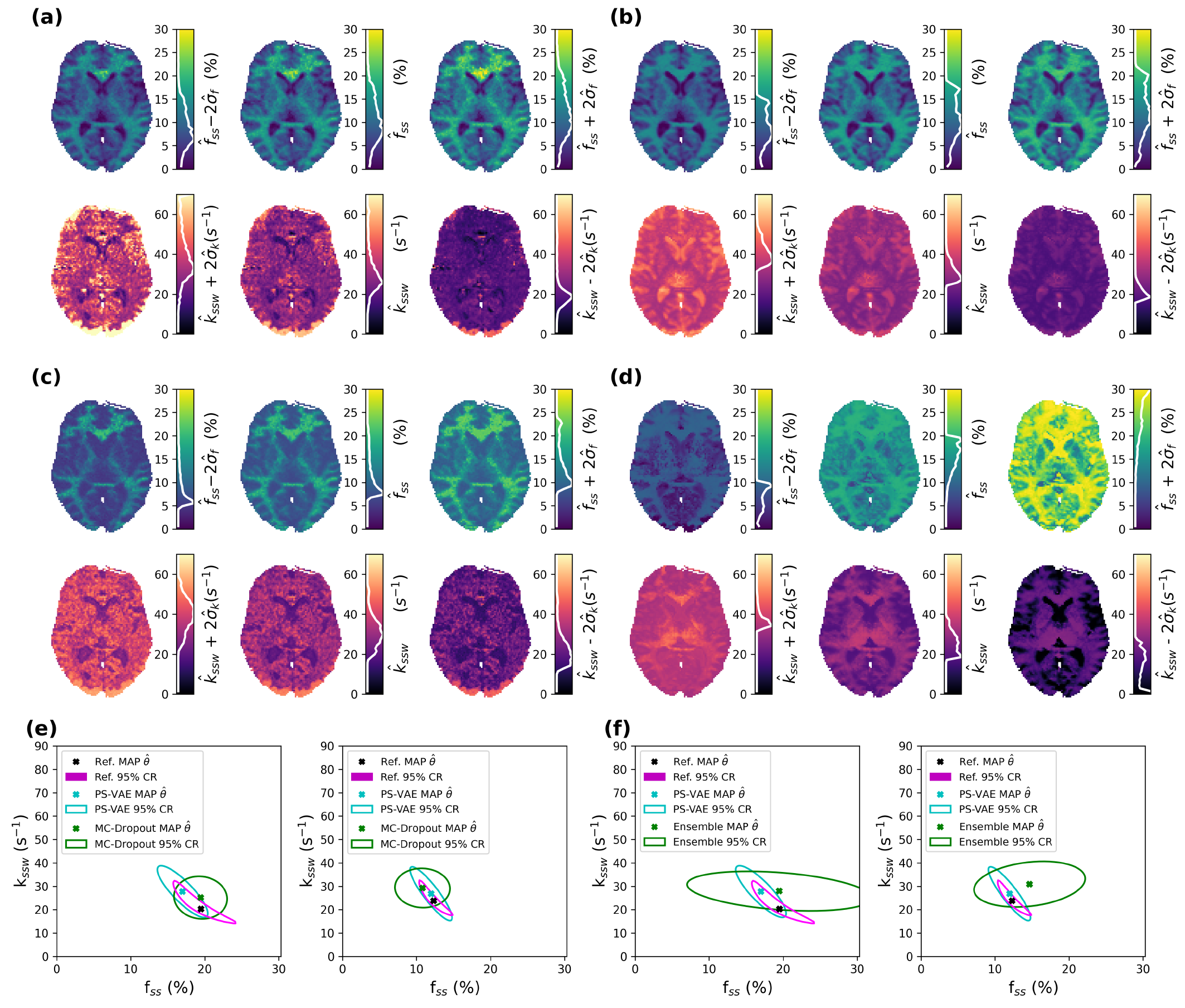} 
\caption{\revised{\textbf{Ablation study and comparison between the proposed PS-VAE and alternative methods using semisolid MT data from a human volunteer}. \textbf{(a-d)}: The color-coded brain images  represent the application of different Bayesian estimation methods: (a) Likelihood mapping ('reference'), (b) PS-VAE, (c) MC-dropout (d) Deep ensemble. \revisedagain{The visualization components follow the conventions established in Fig. \ref{fig:mouse_maps}}
\textbf{(e,f)}: MAP estimates and CRs given by  multiple methods are shown atop the 2D parameter space for two representative voxels (left/right in each panel). The CRs for probabilistic networks (green) are constructed by Gaussian modeling of the results for multiple point-prediction inferences across random dropout realizations or ensemble members (see Fig. \ref{fig:simulated}g,h,j,k for explicit demonstration).
}}
\label{fig:ablations_in_vivo}
\end{figure*}

\revised{
\subsubsection{Comparison with supervised probabilistic networks}
\label{ablation_in_vivo}
While the reference grid-based method used in the previous section may offer a reasonable approximation of the ground truth, it is computationally impractical for whole-brain data (see runtimes provided in Table \ref{table:comp_perf}). To further position the proposed approach within the landscape of NN-based alternative methods, another in vivo comparison study was performed using probabilistic networks trained on synthetic data (Sec. \ref{sec:digital_phantom_design}), in the semisolid-MT PW setup. The representative tissue parameter estimate maps obtained by the reference method (Fig. \ref{fig:ablations_in_vivo}a) were visually more similar to those from PS-VAE (Fig. \ref{fig:ablations_in_vivo}b) than to those from dropout or deep-ensemble alternatives (Fig. \ref{fig:ablations_in_vivo}c,d). The CR analysis of individual voxels (Fig. \ref{fig:ablations_in_vivo}e-f) yet again demonstrated the ability of PS-VAE to capture the correct physics-consistent CR shape and orientation, in contrast to generic alternatives. The typical voxelwise KL-divergence between predicted and reference voxel posterior was quantified across all 3D volunteer scans, yielding 2.7$\pm$0.4 and 4.7$\pm$0.5 for MC-dropout and ensemble, respectively, compared to 2.2$\pm$0.8 for PS-VAE (Table \ref{table:agreement}).
} 

\subsection{Computational performance}
\subsubsection{Benchmarking results}
Complete \revised{run-times are summarized in Table} \ref{table:comp_perf}.
The PS-VAE training for the human 3D brain imaging analysis required 9.5$\pm0.1$ min for the semisolid MT protocol (employing the \textit{isar2} analytical approximation for 2-pool pulsed-saturation BM dynamics) \revised{   and 95.8$\pm$3.4 min for the CEST APT protocol (using the differentiable numerical solution of BM equations under a 3-pool model \cite{roeloffs_towards_2015, finkelstein_multi-parameter_2025})}. The inference of both on a test subject 3D volume was completed in 1.7$\pm$0.05 sec. This is in contrast to the 1.6$\pm$0.05/0.8$\pm$0.05 s (for MT/APT) for a single voxel and 127.4$\pm6.7$ / 64.8$\pm$3.3 min for a slice (projected to an average of 95.5/48.6 hours for the whole brain) required for the reference method (likelihood mapping using 10$^4$ parameter combinations across a 100x100 grid). Fitting in mouse 2D imaging (MT and APT) required 6.4$\pm$0.4 min per mouse (single slice), compared to 0.57$\pm$0.01 s per voxel and 10.0$\pm$1.3 min for a whole slice for likelihood mapping. \revised{Similarly, fitting a semisolid-MT 3D scan of a patient took the same time (9.5$\pm$0.1 min) as training on 3 human volunteer volumes (as the number of steps was fixed).}

\subsubsection{Computational advantages across operation modes}

\revised{
The full acceleration promise of PS-VAE towards real-time Bayesian quantification comes into effect in a "transfer mode" scenario, where a pretrained network is inferred on a newly scanned subject. While this flow was validated in a digital phantom (Fig.\ref{fig:simulated}) and in the 3T scans of healthy volunteers (Fig. \ref{fig:vol_mtcest_maps}), cross-site or unseen-pathology scenarios can present a distribution-shift challenge. The subject-specific learning ('fitting mode', see Sec. \ref{sec:small-data training}) addresses this challenge and has been shown to run in a reasonably short time frame (Table II). This capability is agnostic to pathology and does not rely on any prior training data distribution. Furthermore, it can be adapted to any new scanner, field strength, and protocol by changing the formulation of the forward model used in the ‘test-time training‘ loop. This ‘neural fitting’ approach was demonstrated across the preclinical animal studies (Fig. \ref{fig:mouse_maps}) and in two human subjects with brain cancer and Alzheimer's disease (Fig. \ref{fig:patient_MT}) across two different medical centers. 
}

\begin{table}[t]
\revised{
\centering
\scriptsize
\setlength{\tabcolsep}{3.5pt}
\renewcommand{\arraystretch}{1.15}
\caption{\textbf{Computational performance comparison between PS-VAE and cartesian-grid likelihood mapping (reference method)}}
\label{table:comp_perf}
\begin{tabular}{@{}l l c c@{}}
\toprule
\textbf{Study} & \textbf{Task} & \textbf{PS-VAE} & \textbf{Reference Method}\\
\midrule
Human MT & Training/fitting & 9.5$\pm0.1$ min$^{a}$ & n/a\\
(PW saturation) & Inference (voxel) & 0.14$\pm$0.01 sec & 1.6$\pm$0.05 sec\\ 
& Inference (2D slice) & 0.15$\pm$0.01 sec & 127.4$\pm6.7$ min\\
& Inference (3D volume) & 0.9$\pm$0.03 sec & 95.5$\pm$0.1 hours \\
\midrule
Human APT & Training & 95.8$\pm$3.4 min$^{a}$ & n/a \\
(PW saturation) & Inference (voxel) & 0.21$\pm$0.01 sec & 0.8$\pm$0.05 sec\\
 & Inference (2D slice) & 0.21$\pm$0.01 sec & 64.8$\pm$3.3 min\\
 & Inference (3D volume) &  1.7$\pm$0.05 sec &  48.6$\pm$0.1 hours \\
\midrule
Mouse MT / APT & Fitting (2D slice)   & 6.4$\pm$0.4 min & n/a  \\
(CW saturation) & Inference (voxel) & 0.2$\pm$0.01 sec & 0.57$\pm$0.01 sec \\
             & Total$^b$ (2D  slice)    & 6.4$\pm$0.4 min & 10.0$\pm$1.3 min \\
\end{tabular}

\flushleft{$^a$Training on 3D data from three subjects or fitting a patient (5000 steps in either case). $^b$For the mouse brains, PS-VAE operates in a “subject-specific training” (fitting) mode; therefore, the total time to obtain posteriors for all voxels in a 2D slice includes the fitting step (with inference embedded). CW - continuous wave. PW - pulsed wave. 
}
}
\end{table}

\begin{figure*}[!t]
\centering
\includegraphics[width=0.95\textwidth]{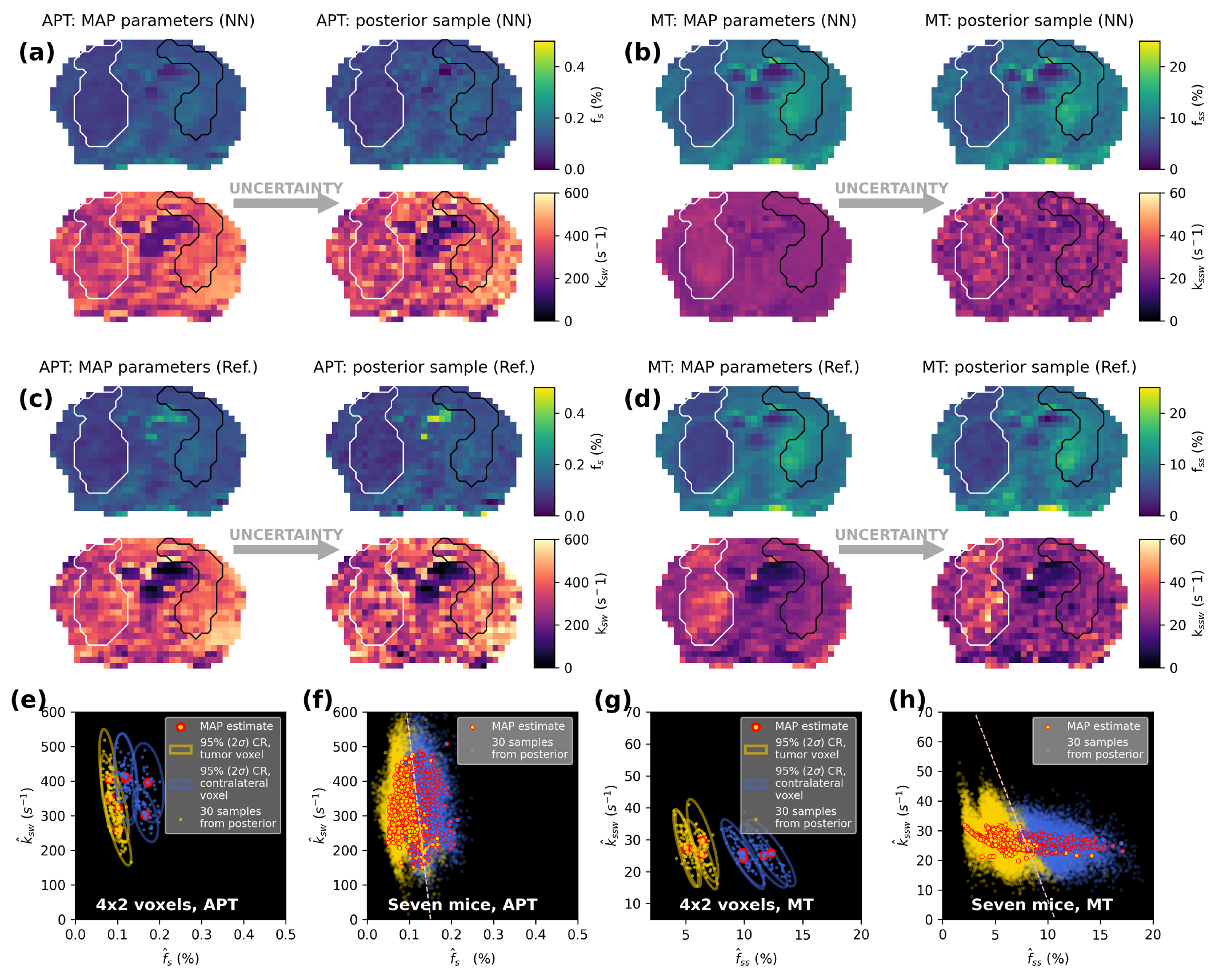} 
\caption{\textbf{Posterior sampling as a means for mitigating denoising priors and evaluating multi-parameter biomarker potential.} \textbf{(a,c)} Quantitative amide proton volume fraction (f$_{s}$, top) and exchange rate (k$_{sw}$, bottom) in a tumor-bearing mouse, obtained using PS-VAE-based point estimates (a, left), the reference method (c, left), and by random sampling of corresponding voxelwise posteriors (a,c, right). 
\textbf{(b,d)}. Quantitative semisolid MT proton volume fraction (f$_{ss}$, top) and exchange rate (k$_{ssw}$, bottom) in the same mouse, obtained using PS-VAE-based point estimates (b, left), the reference method (d, left), and by random sampling of PS-VAE-based voxelwise posteriors (b,d, right). Tumor and contralateral ROIs (as used for voxel classes in e-h) are shown with white and black contours respectively. See \textbf{Supplementary Video 1} for an animated visualization of voxelwise posterior sampling variability across subjects. \textbf{(e-h)} Uncertainty-aware assessment of a multi-parametric quantitative biomarker potential.  \textbf{e,g}. A demonstration of the posterior sampling concept across 8 voxels (color-coded by tissue type, 4 for each), showing the PS-VAE MAP estimates (red-circled points), the bounds of 95\%-CRs obtained from the PS-VAE posteriors, and 30 individual samples per voxel (non-circled points) for APT (e) and semisolid MT (g). \textbf{f,h}. A UQ-informed linear tissue classifier (dashed white line) is obtained by training on data augmented by posterior sampling, following pixel 
aggregation from multiple subjects for APT (f) and semisolid MT (h) imaging. 
Note the agreement between methods as well as between spatial/posterior variation. The spatial-smoothing effect of neural estimation is seen for k$_{ssw}$ (bottom b, g, h), and mitigated by the UQ-based sampling, which protects against over-confidence.
}
\label{fig:classifier_demo}
\end{figure*}

\subsection{Per-voxel Bayesian uncertainty vs. inter-voxel variation}
There are certain pitfalls in applying complex neural networks to parameter-quantification tasks. These include hallucinations of anatomical structures, non-existing contrast, and smoothing. For example, even though we deliberately employed a simplistic network architecture  (voxelwise MLP) to prevent anatomy and brain structure memorization, the NN-based point estimate (MAP) maps for $k_{ssw}$ were still visibly smoothed (reflecting an implicit denoising prior\cite{finkelstein_multi-parameter_2025}). Thus, a simplistic assessment of robustness based on MAP spatial noise may underestimate the true uncertainty in tissue parameter quantification when used with neural estimators. This again highlights the need for UQ. 
While Figs. \ref{fig:mouse_maps}, \ref{fig:vol_mtcest_maps}and \ref{fig:patient_MT} present explicit maps of CI bounds,  Fig. \ref{fig:classifier_demo} represents a further alternative option for visualization, where maximum a posteriori (MAP) estimate maps are compared to a random sampling of voxelwise
posteriors in a tumor-bearing mouse; \textbf{Supplementary Video 1} extends Fig. \ref{fig:classifier_demo}a-d to illustrate the variability across samples and subjects. Notably, the general agreement between the posterior-driven variations and the spatial-based variations provides meaningful validation of the basic assumptions. An exception was observed in the neural estimation of k$_{ssw}$, where the spatial noise appears suppressed (smoothed). However, sampling from the PS-VAE-based posteriors yielded a realistic restoration of the spatial variance (Fig. \ref{fig:classifier_demo}b, bottom), in line with the exact Bayesian reference (Fig. \ref{fig:classifier_demo}d, bottom). Therefore, UQ visualization via direct voxelwise projection of randomly sampled posteriors may serve as another way to mitigate the potential impact of NN biases on the clinical user's perception.
Another implication of our findings, in the context of the search for a quantitative imaging biomarker (QIB) that can be used for tissue classification (e.g., tumor vs normal), relates to the desirability of a "single-number" value that represents the discrimination potential of a given biophysical parameter. Unfortunately, the traditional contrast-to-noise ratio (CNR) metric may be biased when maps are overly smoothed. As an alternative, we propose a probabilistically sound definition of a Contrast-to-total-Uncertainty-Ratio (CUR) for a parameter $\theta_i$ by replacing the denominator calculation in the CNR expression with sampling from voxelwise multi-parametric posteriors $P^{(v)}_{\boldsymbol{\theta}}$:
\begin{equation}
\begin{split}
&CUR(\theta_i,\mathcal{V}_1,\mathcal{V}_2) := \left(\mu_1 - \mu_2\right)/\sqrt{\left(V^{tot}_1+V^{tot}_2\right)/2} 
\\ 
&\mu_{1,2}=\left<\hat{\theta_i}^{(v)}\right>_{v \in \mathcal{V}_{1,2}} 
\hspace{1pt};\hspace{2pt}
V^{tot}_{1,2} = \left< \left(\theta'_i(v)-\mu_{1,2}\right)^2 \right>_{v \in \mathcal{V}_{1,2}}
\\ 
&\theta'_i(v)=(\boldsymbol{\theta} \sim P^{(v)}_{\boldsymbol{\theta}})[i]\
\end{split}
\label{eq:CUR}
\end{equation}
where $\mathcal{V}_{1,2}$ refer to the ROIs of two tissues. Employing this metric on all seven tumor-bearing mice used in this study yielded a CUR of 1.93$\pm$0.50, 0.26$\pm$0.16, 1.23$\pm$0.41, and 0.61$\pm$0.43, for f$_{ss}$, k$_{ssw}$, f$_{s}$, and k$_{sw}$, respectively, in line with the trend revealed by the certainty-based assessment described in Sec. \ref{sec:uncertainty-visualization}.

As shown earlier, PS-VAE inherently extends beyond univariate estimates and may facilitate UQ-aware multi-parametric QIBs\cite{raunig2023multiparametric}. This is demonstrated in Fig. \ref{fig:classifier_demo}f,h, using a simplistic linear 'tissue-classifier', trained using posterior samples aggregated from all mice voxels. Interestingly, the slanted decision boundary in Fig. \ref{fig:classifier_demo}h assigned a role to k$_{ssw}$ in classification, despite its low individual CUR. In other words, while univariate UQ indicates the limited potential of the k$_{ssw}$ parameter as a self-contained tumor biomarker, the joint bivariate UQ suggests that it may still add value when combined with the semisolid MT proton volume fraction, as part of a multi-parametric QIB.

\begin{figure*}[!t]
\centering
\includegraphics[width=0.99\textwidth]{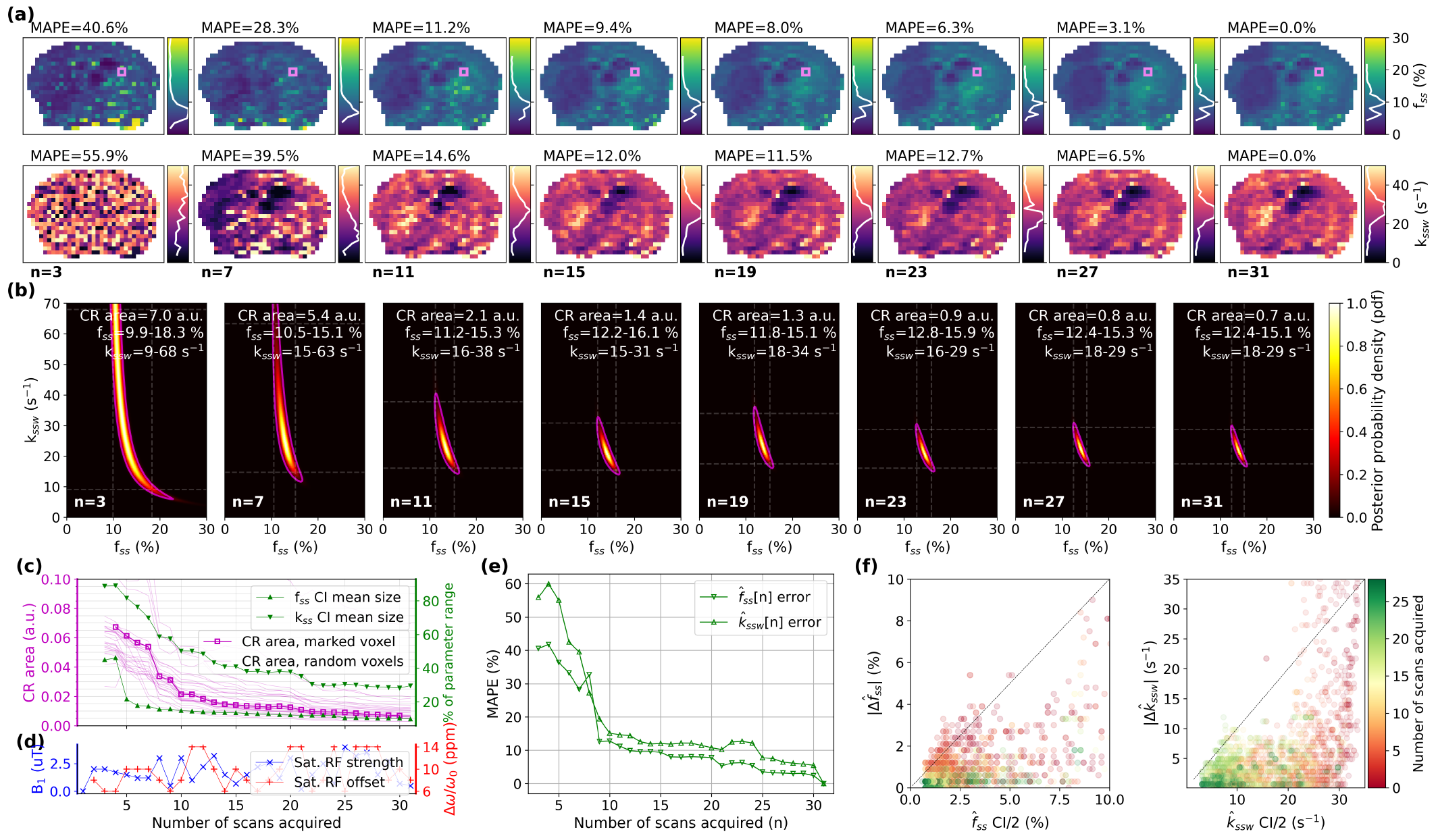} 
\caption{\textbf{Continuous monitoring of multi-parameter posterior dynamics across progressively acquired semisolid magnetization transfer (MT) quantitative images in a tumor-bearing mouse.} (\textbf{a}) Semisolid MT proton volume fraction (f$_{ss}$, top) and exchange rate (k$_{ssw}$, bottom) quantitative estimates, extracted using the dictionary best fit entry. The retrospective early-stopping error ($\Delta\theta_n=|\hat{\theta}(S_{[1,..,n<N]})-\hat{\theta}(S_{[1,..,N]})|; \theta \in \{f,k\}$, aggregated over voxels as mean absolute percent error (MAPE): $\overline{\Delta\theta/\theta}$ (\%)) 
is presented above each dynamic image. (\textbf{b}) Bivariate exact posteriors (heatmap)  and confidence region (CR) bounds (magenta) derived by Eqs. \ref{Eq:s_m}-\ref{eq:err2prob}, for a random brain voxel. (\textbf{c}) Calculated CR area and univariate 1D confidence interval (CI) widths (green) aggregated over brain voxels. (\textbf{d}) The saturation pulse power and frequency offset used throughout the imaging protocol. (\textbf{e}) The mean relative early-stopping error (MAPE, \%) across partial sequence lengths (n). (\textbf{f}) A comparison between the estimated confidence interval semi-widths (CI/2) and retrospective early-stopping error ($\Delta\theta$) across all brain voxels and n=1..31 acquired images (color-coded).  }
\label{fig:mouse_seq}
\end{figure*}

\begin{figure*}[t]
\centering
\includegraphics[width=0.99\textwidth]{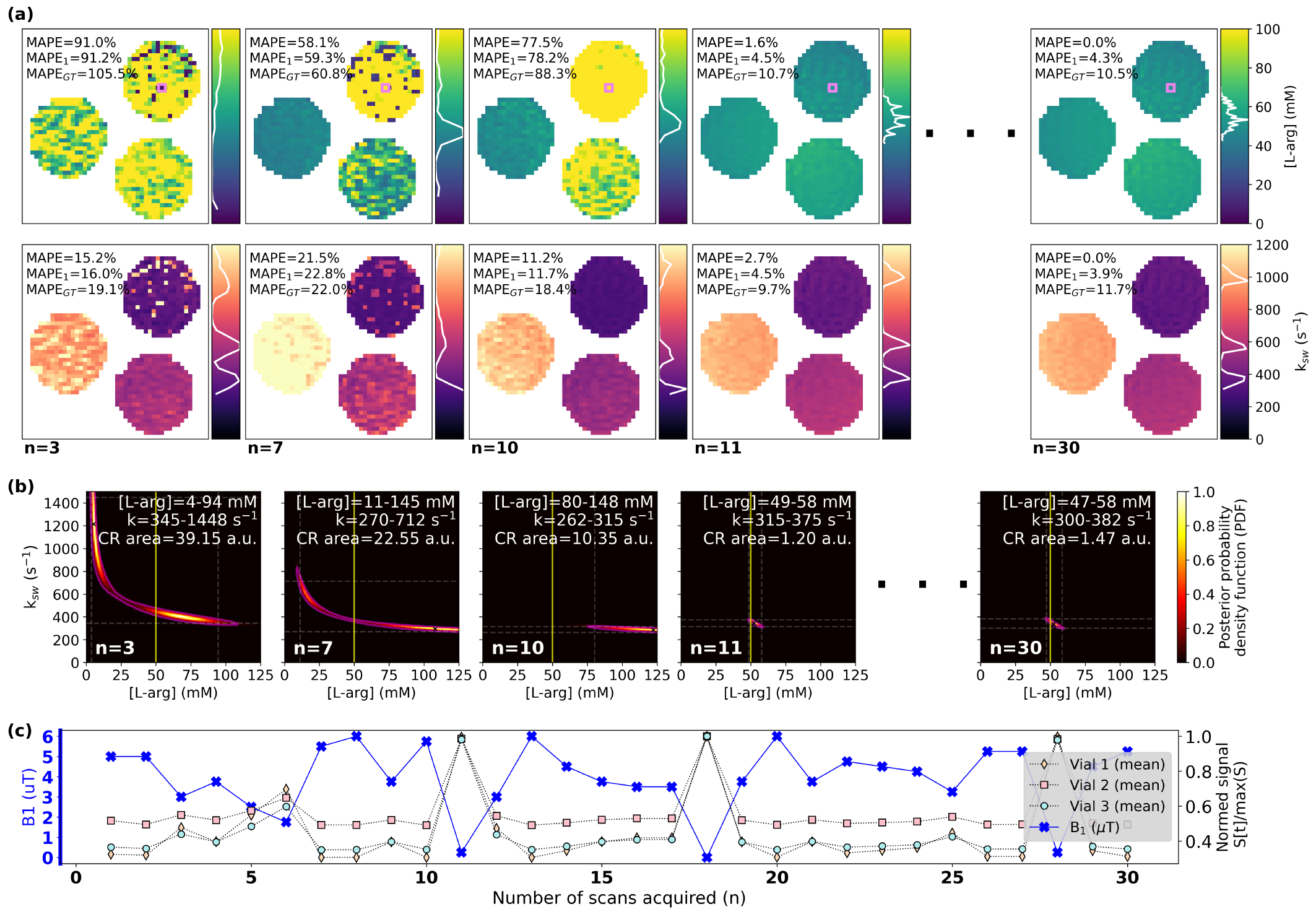}
\caption{\textbf{Continuous monitoring of multi-parameter posterior dynamics across progressively acquired CEST quantitative images in an L-arginine phantom.} (\textbf{a}) L-arginine concentration (top) and proton exchange rate (k$_{sw}$, bottom) quantitative estimates, extracted using the dictionary best fit entry. The early-stopping error (as mean absolute percent error, MAPE) of partial-data estimate ($\hat{\boldsymbol{\theta}}$[n], $\boldsymbol{\theta}:=$\{f$_s$,k$_{sw}$\}) compared voxelwise to the full-protocol (30 images) estimates $\hat{\theta}$[N] is presented for each dynamic image, alongside alternative comparisons: (i) to vial-mean of $\hat{\theta}$ (MAPE$_1$); (ii) to ground truth - known concentrations and proton exchange rate independently estimated using QUESP \cite{cohen_rapid_2018} (MAPE$_{GT}$). (\textbf{b}) Bivariate confidence regions (CRs) for a random phantom voxel. (\textbf{c}) The saturation pulse powers used throughout the imaging protocol (the frequency offset was fixed at 3 ppm \cite{cohen_rapid_2018}); and the resulting signal values (averaged by vial, and normalized by maximal value). 
} 
\label{fig:phantom_seq}
\end{figure*}

\subsection{Adaptive acquisition by multi-parameter posterior tracking} 
The ability to quantify uncertainty rapidly opens new opportunities for online optimization of acquisition protocols, in a subject-specific manner. As a proof of concept, we retrospectively calculated the pixel-wise posteriors for progressively acquired experimental data (applying Eq. \ref{eq:err2prob} with n'$<$n), obtained using CEST (Fig. \ref{fig:phantom_seq}) or semisolid MT (Fig. \ref{fig:mouse_seq}) MRF protocols. Using the posteriors, we tracked CR shrinking as measurements accumulated (Fig. \ref{fig:mouse_seq}b,c and \ref{fig:phantom_seq}b) and used this as a surrogate for quantification accuracy. In other words, the CRs can serve as a direct means for (potentially) informing the MRI operator that the scan can be stopped and no additional data acquisition is needed.  The Pearson's correlations between the CIs and the early-stopping error (aggregated as mean absolute percentage error, MAPE) were (\textit{r}=0.77, \textit{p}$<$0.001) and (\textit{r}=0.95, \textit{p}$<$0.001) for the f$_{ss}$ and k$_{ssw}$ estimation, respectively (Fig. \ref{fig:mouse_seq}e). Furthermore, the width of the CIs provided a tight bound for the quantification accuracy (Fig. \ref{fig:mouse_seq}f), as the de-facto errors of f$_{ss}$ and k$_{ssw}$ estimation due to early stopping were consistently smaller than the CI-based bound - in 96.5\% and 94.3\% of the samples, respectively. In both animal and phantom studies, convergence to stable CRs/CIs, and, in turn, to a stable quantification error, occurred before 50\% of the original sequence was completed, suggesting that substantial acceleration can be achieved with only a minor degradation in performance. The CR convergence rate was generally in line with the diversity of saturation pulse parameters offered by the acquisition protocol, as most intuitively shown at n=11 for the phantom imaging case (Fig. \ref{fig:phantom_seq}c). Adding an additional measurement with a B$_1$ distinctly different from previous measurements yielded improved encoding capability. This finding is consistent with previous protocol optimization reports \cite{perlman_cest_2020} that relied solely on comparing quantification maps (Fig. \ref{fig:phantom_seq}a) with the final output (i.e., the full acquisition protocol length). In contrast, here we offer a roadmap to real-time adaptivity through CR-tracking (Fig. \ref{fig:phantom_seq}b).

\revised{
\subsection{Limitations and future work} 
\subsubsection{Failure case analysis}
\label{sec:failure_cases}
Several failure modes and degraded performance cases were detected throughout this work:
(i) A canonical failure mode happens in the CSF voxels (Fig. \ref{fig:vol_mtcest_maps} f,m) where the signal’s negligible deviation from a water-only model (no APT pool contribution) translates into a substantial ambiguity. As the measured signal is related to the product of the proton volume fraction and exchange rate \cite{zaiss2022theory},  both f$\sim$0 and k$\sim$0 solutions can be obtained. While the grid-based Bayesian method captures the ambiguity as a posterior bounded by the axes and a curve resembling a $fk=\epsilon$ hyperbola (Fig. \ref{fig:vol_mtcest_maps} f,m), the PS-VAE reveals a breakdown of the Gaussian approximation. However, a user observing such an elliptical PS-VAE CR that simultaneously crosses both the f and k axes could infer that the model's quantitative output is untrustworthy and that the qualitative takeaway at that voxel is ‘no APT pool detected’.\\
(ii) When considering the various combinations of saturation pulse modes and target compounds used in this study, the PS-VAE posteriors associated with the PW-acquired APT data provided the lowest performance (Table \ref{table:agreement}). While one could relate this finding to the generally lower SNR of APT compared to semisolid MT data (associated with an order of magnitude lower proton volume fraction), the tumor-bearing mouse data suggests otherwise, as in that case, the APT yielded better divergence metrics compared to the semisolid MT (1.19±0.53,  0.26±0.07, and 0.44±0.06 vs 1.74±0.23, 0.45±0.06, and 0.61±0.05, for the Kullback-Leibler divergence, Jensen-Shannon divergence, and Hellinger distance for the APT and semisolid MT posteriors, respectively, Table \ref{table:agreement}). Therefore, we postulate that it is the combination of 
the magnitude of the effect with the more complex dynamics of the spin-locked PW saturation pulses that underlies the observed performance degradation. Notably, some groups and sites use an interleaved RF sources approach, which enables pseudo-CW saturation on clinical 3T scanners, potentially mitigating this limitation \cite{zhou2022review}. As the obtained posteriors also depend on the parameter-encoding capability of the specific pulse sequence used, future optimization efforts on these sequences could further improve performance \cite{vladimirov2026optimization}.

\subsubsection{Signal modeling imperfections and potential biases}
The sources of fitting discrepancy are aleatoric (noise associated with a given acquisition protocol) and epistemic (model mismatch and auxiliary-parameter error). Inherent contributing factors are the parameter coupling (e.g., between the proton exchange rate and volume fraction) and confounding proton pools that directly affect the tissue parameter discrimination ability. In any quantitative ST experiment, a variety of assumptions are naturally made to somewhat reduce the highly multi-dimensional problem complexity, such as enforcing explicit parameter limits and a specific signal lineshape. For instance, in various studies (including this one), the semisolid MT T$_1$ is assumed to be equal to the water T$_1$ or assigned a fixed value (e.g., 1 sec). Notably, there is increasing evidence of both much shorter T$_1$ for semisolid macromolecules compared to water, and the need to incorporate exchange in modeling of relaxation \cite{asslander2024unconstrained, asslander2026sensitivity, van2016effects, helms2009vivo}. This raises concerns about substantial misspecification, which could fundamentally limit the ability to quantify; the data analyzed in this work, however, did not show signs of severe breakdown, misidentification, or multi-modal degeneracy, with a single notable exception for CSF voxels where both a close-to-zero proton exchange rate and close-to-zero proton volume fraction equally well fit the data, as expected (see Fig. \ref{fig:vol_mtcest_maps}f,m, f,m). Typical deviations increase the signal discrepancy and in parallel, the uncertainty, preserving appropriate coverage. In that sense, Fig. \ref{fig:simulated}a demonstrated the robustness of the PS-VAE UQ coverage for modeling misspecification in a controlled setup. However, biases of the estimates and their CRs due to inaccurate model assumptions are not currently precluded, and future work is needed to better establish such reliability in vivo, where a definite ground truth is lacking.
In addition, similar to previous quantitative ST reports \cite{vladimirov2025quantitative}, our study used separately measured T$_1$ and T$_2$ maps to reduce the parameter space complexity. However, such measurements are inherently affected by non-water proton pools. One potential way to address this limitation in the future is to expand the current modeling to simultaneously extract water tissue parameters, along with semisolid MT and CEST proton exchange parameters, using all available raw measurements. Applying PS-VAE to this higher-dimensional case may involve technical challenges yet could ultimately provide even higher speed-ups over non-neural methods while improving robustness.
An alternative approach is to repeatedly iterate between water relaxivity estimation and non-water proton exchange quantification\cite{shmuely2025quantitative}, at each step taking one estimate as fixed while refining the estimation of the other, until reaching convergence.

\subsubsection{Posterior modeling imperfections}
While the general multivariate Gaussian model assumption (as opposed to diagonal-covariance, see Fig. \ref{fig:simulated}c-e) seemed sufficient for approximating the posterior of the ST proton exchange parameters studies in this work (Table \ref{table:agreement}), this may not be the case for other imaging scenarios, e.g. when generalizing to simultaneous Bayesian quantification of a large number of proton pool parameters from all measured data and protocols. Specifically, a multi-modal posterior could arise when assuming multiple proton pools with a close chemical shift. Future work could evaluate whether PS-VAE could address such ambiguities and ill-posedness through 
a specialized encoder. The latter can either output the explicit parameters of a Gaussian mixture model or generate the input-dependent transformations of a conditional normalizing flow \cite{winkler2019learning, denker2021conditional}
towards learning to estimate the density of arbitrarily complex posterior structures, including multimodal ones.
Alternatively, a meaningful prior could be imposed on the physical latent space to enforce domain knowledge onto mode choice, replacing the currently used boxcar (flat over the support) prior.
We note in this context that while the Gaussian approximation can be unreliable when support constraints (set by the boxcar prior) are active, this issue, such as the k$_{sw}$ upper bound being crossed in Fig. \ref{fig:vol_mtcest_maps}e, did not by itself hinder performance. The only exceptions were when low signal effects occurred alongside an active lower bound at zero (as discussed in Sec. \ref{sec:failure_cases} above).

\subsubsection{Generalization capabilities and clinical utility}
PS-VAE was implemented and validated on phantoms, tumor-bearing mice, healthy human volunteers, and human subjects with Alzheimer's disease or brain cancer across three field strengths and four imaging sites/medical centers. The main motivation for this study design was to investigate the method's compatibility across different imaging scenarios, potentially allowing both preclinical characterization of animal disease models and human studies. As expected, some performance degradation was observed in the human 3T PW experiments compared to animal and phantom CW imaging experiments at higher field strengths, as extensively discussed in Sec. \ref{sec:failure_cases} above. The main difference in performance is due to the greater spectral overlap among proton pools at 3T and the greater complexity of simulating a PW saturation pulse train compared to CW \cite{vladimirov2025quantitative}. Preliminary translation potential was demonstrated in four healthy human volunteers and two subjects with Alzheimer's Disease or brain cancer, who were scanned at three different sites. Nevertheless, a large cohort future study, spanning multiple semisolid MT and CEST targets, needs to be performed to determine the ultimate benefit of using this approach in a clinical setting.}

\section{Conclusion}
We have developed an uncertainty-aware physics-guided quantification framework that provides per-voxel confidence intervals and multivariate confidence region estimates in nearly real time. The results underscore the importance of characterizing joint uncertainty in quantitative molecular MRI.

\bibliographystyle{IEEEtran} 
\bibliography{IEEEabrv,manual,zotero,zotero_old}

\section*{Acknowledgment}
The authors thank Tony Stöcker and Rüdiger Stirnberg for their help with the 3D EPI readout. The study was supported in part by Moonshot-Med, a joint grant program from Tel Aviv University and Clalit Innovation, the innovation arm of Clalit Health Services, and the Edmond J. Safra Center for Bioinformatics at Tel Aviv University. This work was supported by the German Research Foundation (DFG, 559955894) and the Israel Science Foundation (grant No. 1030/25). A.F.
acknowledges the support of the TAD Excellence Program for Doctoral Students in Artificial Intelligence and Data Science. This project was funded by the European Union (ERC, BabyMagnet, project no. 101115639). Views and opinions expressed are, however, those of the authors alone and do not necessarily reflect those of the European Union or the European Research Council. Neither the European Union nor the granting authority can be held responsible for them.


%

\ifCLASSOPTIONcaptionsoff
  \newpage
\fi

\end{document}